\documentclass{article}
\usepackage{graphicx} 

\usepackage{hyperref}
\usepackage{amsmath}
\usepackage{amssymb}
\usepackage{multirow}

\usepackage{tikz}
\usetikzlibrary{matrix}

\usepackage[linesnumbered,ruled,vlined]{algorithm2e}
\usepackage[utf8]{inputenc}

\usepackage{graphicx}
\usepackage{subcaption}

\usepackage{colortbl}
\usepackage{pifont}
\usepackage{array}
\usepackage{multirow}
\usepackage{xcolor}

\usepackage{booktabs} 
\usepackage{threeparttable} 

\usepackage{float}

\usepackage{scrextend}

\usepackage{biblatex}
\bibliography{reference.bib}

\def\email#1{{\texttt{#1}}}
\providecommand{\keywords}[1]
{
  \small	
  \textbf{\textit{Keywords---}} #1
}

\usepackage[title]{appendix}
\usepackage{etoolbox}
\BeforeBeginEnvironment{appendices}{\clearpage}

\title{Electrostatics-based particle sampling and approximate inference}
\author{Yongchao Huang \footnote{Author email: \email{yongchao.huang@abdn.ac.uk}}}
\date{May 2024}

\begin{document}

\maketitle

\begin{abstract}
    A new particle-based sampling and approximate inference method, based on electrostatics and Newton mechanics principles, is introduced with theoretical ground, algorithm design and experimental validation. This method simulates an interacting particle system (IPS) where particles, i.e. the freely-moving negative charges and spatially-fixed positive charges with magnitudes proportional to the target distribution, interact with each other via attraction and repulsion induced by the resulting electric fields described by Poisson's equation. The IPS evolves towards a steady-state where the distribution of negative charges conforms to the target distribution. This physics-inspired method offers deterministic, gradient-free sampling and inference, achieving comparable performance as other particle-based and MCMC methods in benchmark tasks of inferring complex densities, Bayesian logistic regression and dynamical system identification. A discrete-time, discrete-space algorithmic design, readily extendable to continuous time and space, is provided for usage in more general inference problems occurring in probabilistic machine learning scenarios such as Bayesian inference, generative modelling, and beyond.
\end{abstract}

\keywords{Electrostatics; interacting particle system; sampling; approximate inference.}

\section{Introduction}

Probabilistic approaches\cite{Zoubin2015PML,Jordan1999introduction} are used in many learning and decision-making tasks, e.g. Bayesian inference\cite{Winkler1972introduction,Gelman2013BDA} and generative modelling\cite{Murphy2012PML,Bishop2023DL}, and many of them involve estimating the distribution of some quantities of interest. For example, generative adversarial network (GAN\cite{goodfellow2014GAN}), variational autoencoder (VAE\cite{kingma2013VAE}), and more recently diffusion models\cite{scoreSDE2020,Ho2020DDPM}, either explicitly or implicitly learn some distributions of underlying data or variables. However, in many cases the target density is not straightforwardly approachable, e.g. the posterior in a Bayesian model can be analytically intractable \cite{Murray2006Intractable} due to e.g. complex, high-dimensional integration \footnote{Bayesian neural networks (BNNs \cite{Zoubin2016BNN}), for example, issue high-dimensional parameters whose posterior distributions are in general intractable.}, unavailable likelihoods or large data sets \cite{Martin2024}. In such cases, one may resort to approximate inference methods such as Markov chain Monte Carlo (MCMC\cite{MacKay97Intro,Mackay2002Info}) or variational inference (VI) methods\cite{VI_Blei,Fox2012tutorial}. These methods perform inference via sampling or optimisation routines.

Sampling from a prescribed distribution, either fully or partially known, is not easy. If we know the full probability density function (\textit{pdf}), several statistical methods are available, varying in the resulting sample quality though, for producing (quasi) samples. These include inverse transform sampling \cite{Devroye1986}, Box-Muller transform \cite{Box1958}, Latin hypercube sampling (LHS) \cite{McKay1979}, rejection sampling \cite{vonNeumann1951}, importance sampling \cite{Kahn1949}, Quasi-Monte Carlo Methods (QMC) \cite{Niederreiter1992}, etc. The most flexible approaches, particularly for sampling from partially known densities, are based on Markov chains \cite{Bayesian_signal_processing_Joseph}. For example, MCMC methods such as Metropolis–Hastings (MH) sampling \cite{Metropolis1953}, Gibbs sampling \cite{Geman1984}, Langevin Monte Carlo (LMC) \cite{Roberts1996}, Hamiltonian Monte Carlo (HMC) \cite{Duane1987}, slice sampling \cite{Neal2003slice}, etc. VI methods such as mean-field VI \cite{Jordan1999introduction}, stochastic VI \cite{Hoffman2013}, black-box VI \cite{Ranganath2014}, variational autoencoders (VAEs) \cite{kingma2013VAE}, VI with normalizing flow \cite{Rezende2015}, by its nature are designed for inference and widely used in many applications \cite{Jordan1999introduction,Roberts2002AR}; some particle-based VI (ParVI) methods such as Stein variational gradient descent (SVGD) \cite{Liu2016SVGD}, Sequential Monte Carlo (SMC) \cite{Doucet2001}, particle-based energetic VI (EVI) \cite{Wang2021EVI}, etc, also generate samples for inference. MCMC and VI methods together form the paradigm of approximate inference in the Bayesian context; they differ in that, MCMC draws and uses samples for inference while VI in general casts the inference problem into optimisation. Of these sampling algorithms, some utilise the original pdf or up to a constant (i.e. \textit{gradient-free}), e.g. slice sampling and the MH method, while others, e.g. HMC, LMC and most VI methods, utilise the gradient information to guide sampling (i.e. \textit{gradient-based}). Using only gradients enables sampling and inference from an un-normalised, intractable densities which are frequently encountered in Bayesian inference; however, it also increases computational burden. 

While MCMC is flexible, can handle multi-modal distributions and provides exact samples in the limit, it is computationally intensive and converge slowly due to undesirable autocorrelation between samples \cite{Liu2019VI}, they may also require careful tuning and can suffer from convergence issues. It is often impractical for very large datasets or models with complex structures. VI is deterministic, faster and more scalable, but it can struggle with optimization challenges and introduces approximation bias, and may be limited by the expressiveness of the variational family. Both methods, if not all, require estimating the gradients for sampling or optimisation, which can be expensive. Stochastic gradient estimation can be used but extra care should be taken to reduce variance.

ParVI methods enjoy the key strengths of sampling and VI methods: being fast, accurate, deterministic\footnote{Stochastic gradients can be introduced though, see e.g. \cite{Yang2023OPVI}.}, capable of handling multi-modal and sequential inference. ParVI methods evolve a finite set of particles whose terminal configuration upon convergence approximates the target distribution. The particles are deterministically updated using gradients of an explicit or implicit variational objective (e.g. KL-divergence, ELBO, kernelised Stein discrepancy), yielding flexible and accurate approximation \cite{Liu2019VI}. Compared to MCMC and other VI methods, ParVI offers appealing time-accuracy trade-off \cite{Saeedi2017DPVI}: the non-parametric nature of ParVI gives greater flexibility than classical VIs, while the interaction between finite particles, along with deterministic optimisation, makes them more efficient than MCMC methods. While most currently available ParVI methods require use of gradients, gradient-free ParVI methods have not been explored much. 

Here we introduce a physics-based ParVI method, \textit{EParVI}, which simulates an electrostatics-based interacting particle system (IPS) for sampling and approximate inference. It is deterministic and gradient-free, and has traits of non-parametric flexibility, expressiveness, and approximation accuracy, as well as simplicity. Unlike classic VI methods, it only requires the target density to be queryable up to a constant.

\paragraph{Electrostatic halftoning} 
In physics, electrostatics studies the electromagnetic phenomena and properties of stationary or slow-moving electric charges, i.e. when a static equilibrium \footnote{When there are no moving charges, the particle system is said to have achieved a static equilibrium, i.e. the vector sum of all forces acting upon the particle is zero.} has been established. Digital image halftoning, on the other hand, concerns about the distribution of colors \cite{Furht2008dithering}. Image halftoning algorithms have been developed for sampling problems which often occur in rendering, relighting, printing, object placement, and image visualisation \cite{Schmaltz2010}. Electrostatics-based methods are first introduced to image processing \cite{Schmaltz2010,Gwosdek2014} in which \textit{electrostatic halftoning}, a similar approach to ours, was devised to sample image pixels based on its grey values. Digital images are represented by discrete pixels on a computer (though may not be visually visible) and rendered to finite resolution in tasks such as display and printing \cite{hanson2005halftoning}. This requires (re)creating the illusion of e.g. a grey-scale image by distributing the black dots on a white background (i.e. \textit{digital halftoning} \footnote{We only discuss in the context of digital halftoning which renders gray-scale images (i.e. putting black dots on a white surface), and mainly concern about rendering speed and quality.} \cite{hanson2005halftoning}), or resampling the image to adapt to machine resolution, whilst yielding either energetically or visually optimal results \cite{Schmaltz2010}. To accurately and flexibly represent the original image, Schmaltz et al. \cite{Schmaltz2010} proposed the \textit{electrostatic halftoning} method which follows the physical principles of electrostatics to assign and move negative charges, by interacting with positive charges representing image intensities, to form the distribution of grey values. This method is based on the intuitive assumption that, in a region of constant intensity, black points (represented by negative charges) in the sampled images should be equally distributed, while in other regions, through a process of particle interaction and evolution, their distribution will be proportional to the image grey values. 

We therefore note the similarity between (standard and quasi) Monte Carlo sampling and digital halftoning: while both methods aim to resemble a target geometry \footnote{In an inference task, MC methods try to infer the target density; digital halftoning creates a pattern of black dots to replicate the distribution of image intensity.}, they also share the same problem of sample clumpiness \cite{hanson2005halftoning} and empty spots \cite{Ohbuchi1996QMC}, i.e. undesirable clumping of samples (dots in an image) which degrades sampling efficiency or image quality (uneven dot placement that accompanies random dot distributions introduces undesirable texture).

\section{Related work}
Using electrostatics principles to model interactions in a particle system is less exposed but not new. In classic Newton mechanics, the interactions between $M$ bodies is described the Newton's third law $\textbf{F}_i=-\textbf{F}_j, i \neq j, i,j \in \{1,2,...,M\}$, and each individual force follows the Newton's second law $\textbf{F}=m\textbf{a}$ where $m$ is mass and $\textbf{a}$ acceleration. Gravitational force, for example, can be specified as per the inverse-square law: $\textbf{F}=G\frac{m_1 m_2}{r^2}$ where $r$ is the distance between the two interacting objects, and $G$ the gravitational constant. These laws can be derived in a variational form via the energy perspective, i.e. the Maxwell equations (a set of coupled PDEs describing electromagnetic laws), or the Gauss law (which in electrostatics relates the distribution of electric charge to the resulting electric field), the Laplace equation (see Appendix.\ref{app:coulbomb_law_nd}) and the Poisson's equation (generalisation of Laplace's equation) in more specific domains such as mathematics and physics. Regarding a set of charges as particles and modelling their interactions using already-known electrostatic principles requires an alignment between the the behaviors of the particle system we aim to model and that of the charges. In particle-based variational inference (ParVI), the SVGD method \cite{Liu2016SVGD} for example, particles interact and dynamically move towards forming the target geometry; in the charge system, positive and negative charges attract and repel each other and achieve certain distribution at equilibrium. Although we don't intend to binarily classify particles in ParVI (which is essential in electrostatics), the similarity between both inspires the design of new algorithms at their intersection, by taking advantages of both worlds.

A straightforward application of the interacting electrostatic particle system emerged in digital image processing \cite{Ohbuchi1996QMC,hanson2005halftoning,Schmaltz2010}. Inspired by principles of electrostatics, Schmaltz et al. \cite{Schmaltz2010} proposed the \textit{electrostatic halftoning} method for image dithering, stippling, screening and sampling. This physics-based image processing method is simple, intuitive and flexible: inkblots are modelled as small, massless particles with negative unit charge, while the underlying image is modelled as a positive charge density, the attracting forces attract negative charges move proportionally towards dark image regions. Eventually, the particle system reaches an equilibrium state that forms an accurate halftone representation of the original image. This method can deal with both continuous and discrete spaces, and works well on both color and gray-scale images. It also enables arbitrary number of additional samples to be added to an existing set of samples. However, their vanilla implementation has a runtime complexity of $\mathcal{O}(M^2)$ in the number of particles $M$, making it impractical for real-world applications (e.g. high-resolution image halftoning). Gwosdek et al. \cite{Gwosdek2014} proposed a GPU-parallel algorithm to improve the computational efficiency of the electrostatic halftoning method. They simplified the pairwise force calculation by evaluating the near-field and far-field interactions using different strategies, e.g. a novel nearest-neighbour identification scheme was designed, as well as a GPU-based, non-equispaced fast Fourier transform (NFFT) algorithm devised for evaluating interactions between distant particles, which reduces the runtime complexity to $\mathcal{O}(M \log M)$.

Albeit similarities exist between MC sampling and halftoning patterns, the intersection between MC and halftoning techniques has not received much attention in the past \cite{Ohbuchi1996QMC,hanson2005halftoning}. Ohbuchi et al. \cite{Ohbuchi1996QMC} applied the QMC method, which employs deterministic low-discrepancy sequences (LDSs, a screen pixel sampling method), to solve the global illumination problem in 3D computer graphics. They used QMC sequences to improve rendering of surfaces and shadows in a 3D scene, and showed that, QMC integral with LDSs converges significantly faster than MC integral and yielded more accurate results. This QMC-LDS method also allows for incremental sampling. 

Inspired by halftoning, Hanson \cite{hanson2005halftoning} designed a Quasi-Monte Carlo (QMC) method for generating random points sets for evaluating 2D integrals, which reduces the number of function evaluations and yields more accurate estimates than standard QMC methods. Although computation is cheap, however, this method is inapplicable to dimensions larger than four or five as it relies on the calculation of the dimension-restricted mean-squared error. This haltoning-inspired QMC method can also be applicable to sensitivity analysis of computer models. They also proposed a potential-field approach, which considers the repulsion between particles as well as between particles and the boundaries, to obtain 2D random, quasi-uniform integration points.

\section{Main contributions}

We re-invent the idea of electrostatic halftoning \cite{Schmaltz2010} for modelling the interacting particle system (IPS), and apply it to sample and infer probability densities. Although the physics is simple, and the idea of modelling IPS using electrostatics already exists, this work is novel in following three aspects: (1) it is for the first time principles of electrostatics and Newton mechanics are introduced to variational inference; (2) simplified IPS modelling; (3) we extend its theoretical applicability into arbitrary dimensions. The repulsion-attraction framework is different from previous work (e.g. \cite{Schmaltz2010}) in that, we generalise the detailed physics to any dimension, and it is more flexible to set the parameters, e.g. charge magnitudes with an annealing scheme, by relaxing the equilibrium condition for accelerating convergence. We also introduce two new particle updating rules, rather than assuming linear relation between force and displacement, for yielding more accurate approximation. Other tricks such as particle filtering, noise perturbation and probabilistic move are also introduced.

The paper is organised as follows: Section.\ref{sec:methodology} illustrates the methodology; experimental validation follows in Section.\ref{sec:experiments}; discussion, conclusion and future work are presented in Section.\ref{sec:discussions} and Section.\ref{sec:conclusion_and_future}. Interested readers are referred to Appendicies for more details.

\section{Electrostatics-based sampling} \label{sec:methodology}

\begin{figure}[ht]
    \centering
    \includegraphics[width=0.40\columnwidth]{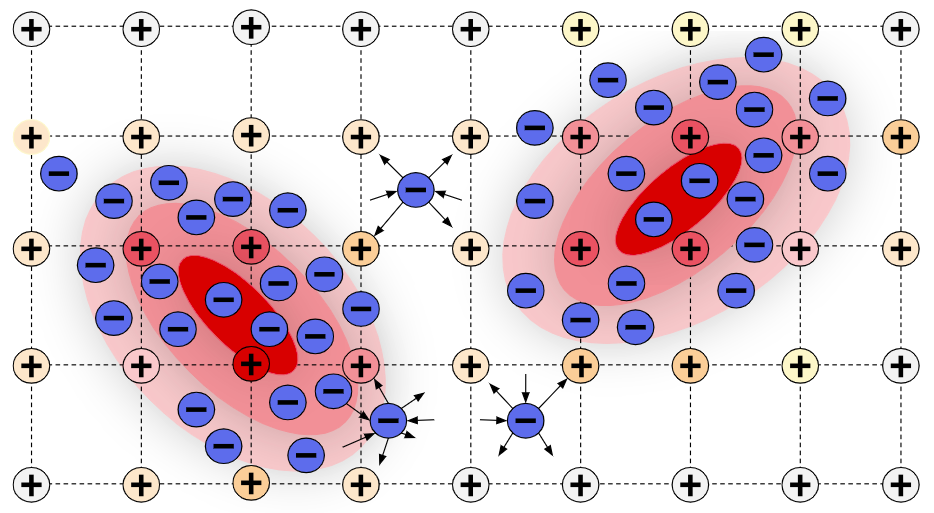}
    \caption{Illustrative diagram of particle distributions. Red darkness indicates the density value of two-Gaussian mixture, arrows denote example forces. Positive charges are fixed at grid points with magnitude proportional to density value, negative charges can move freely.}
    \label{fig:density_diagram}
\end{figure}

The basic idea is simple and rooted in elementary electrostatics: if we place a electric charge, either negative or positive, in a free space without friction \footnote{We restrict the discussion within a free space (i.e. a vacuum), which simplifies the physics without losing generality.}, it generates an electric field which induces attracting or repulsive forces on other charges. If a set of charges of equal value and same sign, all negative for example, are placed, at steady-state these charges will distribute homogeneously and pairwise distances are maximized by repulsion. This uniform density generation motivates the design of MC-style integration algorithms \cite{hanson2005halftoning}, and forms a basic assumption for electrostatic halftoning \cite{Schmaltz2010}. If we further place a set of positive charges at fixed locations (e.g. evenly-spaced grids), with pre-assigned magnitudes specified (e.g. proportional to) by a desired distribution (e.g. distribution of grey values in an image) over the space, the initial steady state breaks and a new equilibrium will be established due to repulsion and attraction. Distribution of the negative charges at the new equilibrium, as we shall see later, is exactly the desired distribution. Based on this mechanism, we can assign a set of positive charges at fixed locations with density proportional to a target distribution, and evolve a set of free, negative charges to form the target geometry. The new steady-state distribution of the negative charges is termed the \textit{energetically ideal particle distribution} \cite{Schmaltz2010}. At the new equilibrium, these negative charges serve as samples drawn from the target distribution and can be used to produce, e.g. summary statistics and predictive uncertainties. See Fig.\ref{fig:density_diagram} for an illustration.

The central concept used in our electrostatics-based particle sampling and inference algorithm is the interaction between charges (particles) which is governed by the interactive forces. In the following, we first describe the design principles for particle interactions, i.e. the designation of repulsive and attracting forces. Based on these design, we devise an algorithm for simulating this evolving particle system towards the target distribution. Some background knowledge, along with the detailed derivation process and primary results, e.g. 1D, 2D and 3D electric fields and forces, are presented in Appendix.\ref{app:coulbomb_law_nd}.

\paragraph{Repulsion} To mathematically formalise this idea, we first define the set of charges (interchangeably used with \textit{particles}) $\{q_i|i=1,2,...,M^{neg}\}$, where $M^{neg}$ it the cardinality of this set, in $\textit{d}$-dimensional space, and each is associated with position $\textbf{x}_i$. Consider a single point charge $q_i$, let $\textbf{E}_{q_i}$ be the electric field induced by $q_i$. In electrostatics, a electric field $\textbf{E}$ is proportional to the electric displacement field and equals the negative gradient of the electric potential $\varphi$; the divergence of $\textbf{E}$, which implies the infinitesimal electric flux, is linked to the distribution of a source via the \textit{Poisson's equation} for electrostatics (see Eq.\ref{eq:linear_field}, Eq.\ref{eq:electric_potential}, Eq.\ref{eq:divergence_of_electrical_field} in Appendix.\ref{app:coulbomb_law_nd}). We assume no external source or sink in this charge particle system, which simplifies the Poisson's equation to \textit{Laplace's equation}; solving Laplace's equation gives the potential field $\varphi$, and therefore the electric field $\textbf{E}$ and force $\textbf{F}$ induced by a point charge (see Appendix.\ref{app:coulbomb_law_nd} for details). In $\textit{d}$-dimensional space, the electric field $\textbf{E}_{q_i}(r)$, at a distance of $r$ from a inducing point charge $q_i$ , is derived as:

\begin{equation} \tag{cc.Eq\ref{eq:electric_field_from_potential}}
    \textbf{E}_i (r) = \frac{q_i \Gamma(\frac{d}{2})}{2\pi^{d/2}\varepsilon_0 r^{d-1}} \textbf{e}
\end{equation}

\noindent in which $r$ is the distance from the point charge $q_i$, $\varepsilon_0$ is the the vacuum permittivity constant, $\Gamma(\cdot)$ the \textit{Gamma function}. $\textbf{e}$ is an \textit{unit vector} indicating the direction of the field. 

Imagine we have another negative, point charge $q_j$ locating at a distance of $r_{i,j}$ from $q_i$. To distinguish, we call $q_i$ the \textit{source point} and $q_j$ the \textit{field point}. The electric force exerted by $q_i$ on $q_j$, as per Coulomb's law, is:

\begin{equation} \label{eq:repulsive_force}
\tag{cc.Eq\ref{eq:electric_force_from_potential}}
    \textbf{F}_{i \rightarrow j}^{rep} (r_{i,j}) = q_j \textbf{E}_i (r_{i,j}) =  \frac{q_i q_j \Gamma(\frac{d}{2})}{2\pi^{d/2}\varepsilon_0 r_{i,j}^{d-1}}  \textbf{e}_{i \rightarrow j}
\end{equation}

\noindent where the unit vector $\textbf{e}_{i \rightarrow j}=(\textbf{x}_j-\textbf{x}_i)/r_{i,j}$ gives the out-pointing direction. The arrow sign above implies 'exerting effect on', read from the left object to the right object. According to \textit{Newton's third law}, if we treat $q_j$ as a source point and $q_i$ a field point, $q_i$ experiences the same amount of force with reverse direction $\textbf{e}_{j \rightarrow i} = -\textbf{e}_{i \rightarrow j}$. 

Now we look at the particle set $\{q_i|i=1,2,...,M^{neg}\}$, and consider the combined repulsive force that particle $j$ receives from all other particles, which can be calculated as:

\begin{equation} \label{eq:repulsive_force1}
\tag{Eq\ref{eq:electric_force_from_potential}b}
    \textbf{F}_j^{rep} (\textbf{x}_j) 
    = 
    \sum_{i=1,i \neq j}^{M^{neg}} 
    \frac{q_i q_j \Gamma(\frac{d}{2})}{2\pi^{d/2}\varepsilon_0 r_{i,j}^{d-1}}  \textbf{e}_{i \rightarrow j}
    =
    \sum_{i=1,i \neq j}^{M^{neg}} 
    \frac{q_i q_j \Gamma(\frac{d}{2})}{2\pi^{d/2}\varepsilon_0} \frac{\textbf{x}_j - \textbf{x}_i}{\lVert \textbf{x}_j - \textbf{x}_i \rVert_2^{d/2}}
\end{equation}

\noindent where $\lVert \cdot \rVert_2$ denotes the $L_2$-norm of a vector. When charges (particles) achieve steady-state with uniform distribution, for each particle $j \in \{1,2,...,M^{neg}\}$, the combined repulsive force $\textbf{F}_j^{rep} (\textbf{x}_j)$ becomes zero, as all induced electric fields acting on the particle cancel.

One can conveniently re-write the first half of Eq.\ref{eq:repulsive_force1} in matrix form, by denoting the row vector $\textbf{q}_j=[q_j, q_j,\cdots, q_j]^T$, column vector $\textbf{q}_{1:M^{neg}}=[q_1, q_2,\cdots, q_{M^{neg}}]$, diagonal matrix $\mathbf{\Lambda}=diag[1/r_{1,j}^{d-1}, 1/r_{2,j}^{d-1},...,1/r_{i,j}^{d-1}]$, as follows:

\begin{equation} \label{eq:repulsive_force2}
\tag{Eq\ref{eq:electric_force_from_potential}c}
\textbf{F}_j^{rep} (\textbf{x}_j)
=
\frac{\Gamma(\frac{d}{2})}{2\pi^{d/2}\varepsilon_0}
\textbf{q}_j
\mathbf{\Lambda}
\textbf{q}_{1:M^{neg}}
\end{equation}

\noindent in which $\mathbf{\Lambda}$ can be viewed as some kind of isotropic covariance matrix produced by a kernel (similar to a RBF kernel but not exactly), with each diagonal element measuring a pairwise correlation or inverse-distance between the two point charges $q_j$ and $q_i$. 

\paragraph{Attraction} We have devised the repulsive forces between particles of the same sign (specifically, all negative in our settings), which, under arbitrary initial positions, should generate a uniform distribution of the particles over the free space at equilibrium. To produce an uneven target distribution $p(\textbf{x})$, we need to pull particles towards (neighbourhood) high density region using attracting forces. This can be achieved by introducing a continuous field of positive charges whose magnitudes distribute as, or proportional to, the target density. The induced new electric fields shall attract negative charges towards high probability (indicated by large positive charge magnitudes) regions. Representing a continuous density over the space requires an infinite number of positive charges to be allocated; realistically, we can mesh the space into equal-distance grids, and place large $M^{pos}$ positive charges fixed at these grid points. As one can imagine, the number of positive charges, therefore the mesh granularity, shall impact the smoothness of the inferred density. 

As such, the overall attracting force, resulted from a positive point charge and acting on a negative charge $q_j$, is \cite{Schmaltz2010}: 

\begin{equation} \label{eq:attracting_force}
\tag{Eq\ref{eq:electric_force_from_potential}d}
    \textbf{F}_j^{attr} (\textbf{x}_j) = - \sum_{i'=1,i' \neq j}^{M^{pos}} 
    \frac{q p(\textbf{x}_{i'}) q_j \Gamma(\frac{d}{2})}{2\pi^{d/2}\varepsilon_0 r_{i',j}^{d-1}} \textbf{e}_{i' \rightarrow j}
    =
    - \sum_{i'=1,i' \neq j}^{M^{pos}} 
    \frac{q p(\textbf{x}_{i'}) q_j \Gamma(\frac{d}{2})}{2\pi^{d/2}\varepsilon_0}  \frac{\textbf{x}_j - \textbf{x}_{i'}}{\lVert \textbf{x}_j - \textbf{x}_{i'} \rVert_2^{d/2}}
\end{equation}

\noindent where we have set the positive charge $q_j=q p(\textbf{x}_j)$ with $q$ being a constant scaling factor specifying the relative strength of attracting forces. A negative sign is put ahead as it's attracting force. 

If we ignore the contributing forces from those located far away, in regions with constant probability (and therefore uniform positive charge magnitudes), the attracting forces cancel and repulsion still yields the energetically ideal positions which maximize pairwise distances. In regions with uneven probability distribution, attracting force dominates and binds particles onto locations with higher probability. If mesh grid is set to be infinitely small, $M^{pos}$ grows to infinitely large and the distribution of positive charge magnitude becomes continuous, Eq.\ref{eq:attracting_force} then becomes an integral.

\paragraph{Evolving the particle system} Summation of the repulsive and attracting forces gives the overall force $\textbf{F}_j=\textbf{F}^{rep}_j + \textbf{F}^{attr}_j$, acting on a single free-moving negative charge $j$; this combined force drives the negative charge to move as per \textit{Newton's second law}. At equilibrium (steady-state), the repulsive force equals the attractive force (in magnitude) and $\textbf{F}_j$ vanishes. As we intend to evolve the particle system over time, we introduce the time dimension and denote the time-dependent quantities with superscript $t$. Further, avoiding complex physics, we assume variation of particle displacement $s^t_{q_j}$ is proportional to the corresponding overall acting force $\textbf{F}^t_j$, i.e. $\delta s^t_j \propto \textbf{F}^t_j$, which gives the following position updating rule \cite{Schmaltz2010} for the particle $j$:

\begin{equation} \label{eq:linear_position_update}
\begin{aligned}
    \textbf{x}^{t+1}_j &= \textbf{x}^{t}_j + \tau \textbf{F}^t_j \\
    &= \textbf{x}^{t}_j + \tau [
    \sum_{i=1,i \neq j}^{M^{neg}} 
    \frac{q_i q_j \Gamma(\frac{d}{2})}{2\pi^{d/2}\varepsilon_0} \frac{\textbf{x}_j - \textbf{x}_i}{\lVert \textbf{x}_j - \textbf{x}_i \rVert_2^{d/2}}
    - \sum_{i'=1,i' \neq j}^{M^{pos}} 
    \frac{qp(\textbf{x}_{i'}) q_j \Gamma(\frac{d}{2})}{2\pi^{d/2}\varepsilon_0}  \frac{\textbf{x}_j - \textbf{x}_{i'}}{\lVert \textbf{x}_j - \textbf{x}_{i'} \rVert_2^{d/2}}
    ] \\
\end{aligned}
\end{equation}

\noindent where $\tau$ is a constant similar to the \textit{compliance} constant in a material \footnote{The compliance is the reciprocal of elastic modulus - both describe the stress-strain behaviour of a material.}; it is suggested to be small to well approximate a linear relation between force and displacement. Note the index $i \in \{1,2,...M^{neg}\}$ indexes the negative charge group, while $i' \in \{1,2,...M^{pos}\}$ is for the positive charge group, and if computation allows, the positive charge group can be made continuous over the space.

Assuming a linear relation between force and displacement violates Newton's second law $\textbf{F}=m\textbf{a}$ where $\textbf{F}$ is the overall acting force, $m$ the mass and $\textbf{a}$ the acceleration; however, this linearisation simplifies and accelerates calculation. Theoretically, to recover Newton's second law, the proportionality constant $\tau$ should be non-linear and time-variant. Here we take another approach to (approximately) preserve this physical law using finite difference. We know the following central difference approximation holds (we drop the index $j$ for now):

\begin{equation} \label{eq:central_finite_diff}
    \textbf{a} = \frac{d^2 \textbf{x}}{dt^2} \approx \frac{\textbf{x}^{t+1}-2\textbf{x}^t+\textbf{x}^{t-1}}{\Delta t^2}
\end{equation}

\noindent as per Newton's second law, we have $\textbf{a}=\textbf{F}/m$, and if we set $m=1$, we obtain $\textbf{F}=\textbf{a}$. We directly have:

\begin{equation} \label{eq:nonlinear_position_update1}
\tag{\ref{eq:linear_position_update}b}
    \textbf{x}^{t+1}_j \approx \textbf{a} \Delta t^2 + 2\textbf{x}^t_j - \textbf{x}^{t-1}_j = \textbf{F}_j \Delta t^2 + 2\textbf{x}^t_j - \textbf{x}^{t-1}_j = \textbf{x}^t_j + \textbf{F}_j \Delta t^2 + \Delta \textbf{x}^{t-1}_j
\end{equation}

\noindent with $\Delta \textbf{x}^t_j=\textbf{F}_j \Delta t^2 + \Delta \textbf{x}^{t-1}_j$. We can also design a damped version with a discount factor $\tau'$:

\begin{equation} \label{eq:nonlinear_position_update2}
\tag{\ref{eq:linear_position_update}c}
    \textbf{x}^{t+1}_j = \textbf{x}^{t}_j + \tau' \Delta \textbf{x}^t_j = \textbf{x}^{t}_j + \tau' [\textbf{F}^t_j \Delta t^2 + \Delta \textbf{x}^{t-1}_j]
\end{equation}

\noindent where $\textbf{F}^t_j$ remains the same as in Eq.\ref{eq:linear_position_update}. Both Eq.\ref{eq:nonlinear_position_update1} and Eq.\ref{eq:nonlinear_position_update2} require $\Delta t^2$ to be sufficiently small to be accurate. The newly introduced hyper-parameter $\Delta t^2$ balances the relative importance of the contribution from current step force $\textbf{F}^t_j$ and the historical displacement $\Delta \textbf{x}^{t-1}_j$ carried over from previous step. This momentum-like updates, hopefully, will give higher fidelity in suggesting next particle position. Note the time discretisation size $\Delta t$, along with the step size $\tau'$, plays a role in determining the accuracy and speed of convergence.

Now we have 3 rules, i.e. Eq.\ref{eq:linear_position_update}, Eq.\ref{eq:nonlinear_position_update1} and Eq.\ref{eq:nonlinear_position_update2}, for updating the negative charge positions in any iteration. We refer to the linear mapping between force and displacement in Eq.\ref{eq:linear_position_update} as the \textit{simple Euler method}, the non-linear approximations in Eq.\ref{eq:nonlinear_position_update1} as the \textit{Verlet integration} or \textit{leapfrog method}, and Eq.\ref{eq:nonlinear_position_update2} as the \textit{damped Verlet integration} method. Simple Euler method may require very small time steps to converge accurately, which leads to slow simulation. The Verlet integration method is commonly used in molecular dynamics. It takes into account the previous position, potentially leading to smoother trajectories \footnote{It also provides better energy conservation and handles oscillatory systems more effectively.}; however, it requires more storage as well. The damped Verlet integration provides additional damping but also has more parameter to tune.

Eq.1s define a set of \textit{discrete-time}, \textit{discrete-space} formulas for updating particle positions. In order to estimate next particle position, we only need to record the historical positions $\{\textbf{x}^t,t=1,2,...,k\}$ and calculate the current time overall force $\textbf{F}_j^t$. We keep updating particles' positions until converge (indicated by either static or small movements). When attaining equilibrium, we have $\textbf{F}^t_j=0$ and $\Delta \textbf{x}^{t-1}_j=0$ for all $j \in \{1,2,...,M^{neg}\}$. The charge system then becomes electrically neutral. As the number of negative particles attracted to a region is proportional to the magnitudes of nearby positive charges, particle clustering at the steady state shall align with the distribution of the magnitudes and therefore the target distribution. A vanilla implementation of the \textit{EParVI} algorithm is described in Algo.\ref{algo:EParVI}.

\begin{algorithm}[ht]
\caption{EParVI}
\label{algo:EParVI}
\begin{itemize}
\item{\textbf{Inputs}: a queryable target density $p(\textbf{x})$, particle dimension $m$, initial proposal distribution $p^0(\textbf{x})$, total number of iterations $T$, total number of negative $M^{neg}$ and positive $M^{pos}=M^m$ charges (where $M$ is the number of positive charges along each dimension), Euler proportionality constant $\tau$, max positive charge magnitude $q$.}  
\item{\textbf{Outputs}: Particles $\{\textbf{x}_i \in \mathbb{R}^m \}_{i=1}^{M^{neg}}$ whose empirical distribution approximates the target density $p(\textbf{x})$.} 
\end{itemize}
\vskip 0.06in
1. \textbf{\textit{Initialise}} particles. \\
    \begin{addmargin}[1em]{0em}%
    (1) Mesh the space into $M^{pos}=M^m$ equidistant grids. Query (and optionally normalise) the density values $p(\textbf{x}_i)$ at all grid points $\textbf{x}_i, i \in \{1,2,...,M^{pos}\}$. \\
    (2) Assign a positive charge with magnitude $qp(\textbf{x}_i)$ to each grid point $\textbf{x}_i$.  \\
    (2) Draw $M^{neg}$ $m$-dimensional negative charges $\textbf{x}^0=\{\textbf{x}_j^0\}_{j=1}^{M^{neg}}$ from the initial proposal distribution $p^0(\textbf{x})$. \\ 
    \end{addmargin}

2. \textbf{\textit{Update}} negative charge positions. \\
For each iteration $t=0,1,2,...,T-1$, repeat until converge: 
    \begin{addmargin}[1em]{0em}%
    (1) Calculate the overall force acting on each negative charge $j$ (see Eq.\ref{eq:repulsive_force1} and Eq.\ref{eq:attracting_force}):  
        $\textbf{F}_j^t = \textbf{F}_j^{rep,t} + \textbf{F}_j^{attr,t}$ \\
    (2) Update negative charge positions (Eq.\ref{eq:linear_position_update} as an example):
        $\textbf{x}^{t+1}_j = \textbf{x}^{t}_j + \tau \textbf{F}^t_j$ \\
    \end{addmargin}

3. \textbf{\textit{Return}} the final configuration of negative charges $\{\textbf{x}_j^{T}\}_{j=1}^{M^{neg}}$ and their histogram and/or \textit{KDE} estimate for each dimension. \\
\end{algorithm}

The \textit{EParVI} algorithm is particularly advantageous for sampling and inferring a partially known density $p(\textbf{x})=p'(\textbf{x})/Z$, e.g. an energy model $p(\textbf{x}) = e^{-f(\textbf{x})}/Z$ where the normalising constant \footnote{$Z$ is also called \textit{evidence} or \textit{marginal distribution} in Bayesian inference. It is generally intractable in high-dimensions due to complex integral.} $Z$ is unknown: we can use this algorithm to effectively draw samples from $p(\textbf{x})$ and make inference about its statistics (e.g. multi-modal means and variances), once the numerator is queryable; we only need to set the positive charge magnitudes $qp(\textbf{x})$ proportional to the queryable term and possibly normalise them. Therefore, in this context we don't distinguish the case where the target density is known up to a constant - we use the same, overloaded notation $p(\textbf{x})$ for denoting either the queryable term $p'(\textbf{x})$ or the full density.

In our experiments, for simplicity, we use equally-charged unit negative charges, and make the maximum magnitude for positive charges to be unity as well, i.e. $q_i=q=1$. Note in this setting, the absolute value of $q_i$ or $q$ is not relevant; they are absorbed into the hyper-parameters $\tau$ and/or $\Delta t^2$. However, one does have the option to set the positive charge constant $q$ to be multiples of the negative charge magnitude $q_i$ to address the relative importance of attracting force. Also, if the force is too large, the negative charge can jump to and get trapped in a low-density region (as discussed in Section.\ref{sec:discussions}), this can be avoided by normalising the force in each iteration. Besides, extra force terms can be added to $\textbf{F}_j^t$, e.g. a dragging force controlling the negative charge to move within certain grids \cite{Schmaltz2010}, which is similar to regularisation used in many objectives. Also, the move $\textbf{x}^{t+1}_j = \textbf{x}^{t}_j + \tau \Delta \textbf{x}^t_j$ can be made probabilistically, i.e. we can insert an MH step to assess if this move is worthwhile. These extensions are discussed in Section.\ref{sec:discussions}.

\paragraph{Computational complexity}
EParVI avoids the expensive computation of gradients; however, it also performs many target function evaluations. Magnitudes of the positive charges at grid points can be pre-calculated and cached for use in all iterations, then in each iteration, we only calculate the pairwise distance between negative and positive charges and query the corresponding magnitudes of positive charges. In each iteration, computing the pairwise forces takes order $\mathcal{O}(m(M^{neg}+M^{pos})^2)$ \textit{FLOP}s, where the number of positive charges $M^{pos}=M^m$, with $M$ being the number of equidistant grid points in each dimension, grows exponentially with the dimension $m$. The overall computational cost over $T$ iterations is therefore $\mathcal{O}(Tm(M^{neg}+M^{pos})^2)$. Major computational efforts can be saved by a wise routine, as proposed in \cite{Gwosdek2014}, which reduces the efforts for computing the pairwise forces to $(M^{neg}+M^{pos}) \log (M^{neg}+M^{pos})$. Of course, one can reduces the total number of positive charges $M^{pos}$ by designing an irregular grid for assigning the positive charges. As discussed in Appendix.\ref{app:note_on_complexity}, once can also save some computational efforts by performing neighbourhood search which identifies near-field particle sets via e.g. \textit{k-d-tree} search, drops contributions from distant particles and reduces the computation, as well as using neural approximation of the force field.

\section{Experiments} \label{sec:experiments}

\subsection{Toy examples via EParVI}

\begin{figure}[ht]
    \centering
    \begin{minipage}[b]{1.0\columnwidth}
        \centering
        \includegraphics[width=0.15\columnwidth]{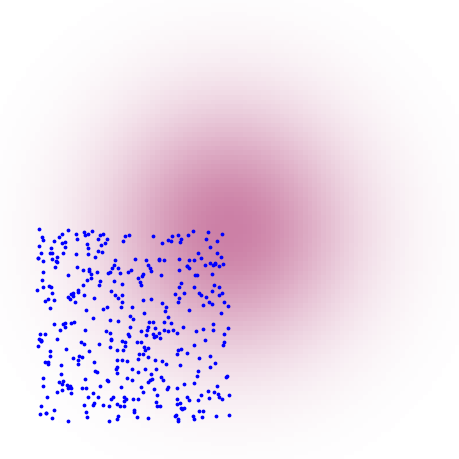}
        \includegraphics[width=0.15\columnwidth]{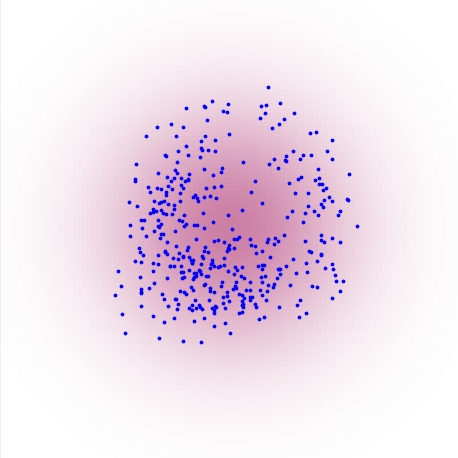}
        \includegraphics[width=0.15\columnwidth]{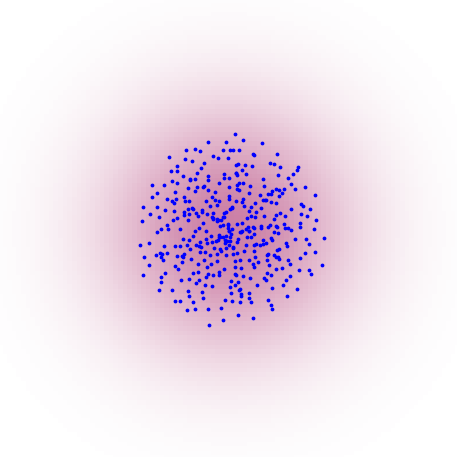}
        \includegraphics[width=0.15\columnwidth]{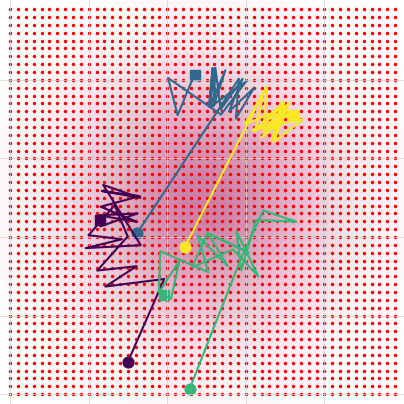}
    \end{minipage}
    
    \begin{minipage}[b]{1.0\columnwidth}
        \centering
        \includegraphics[width=0.15\columnwidth]{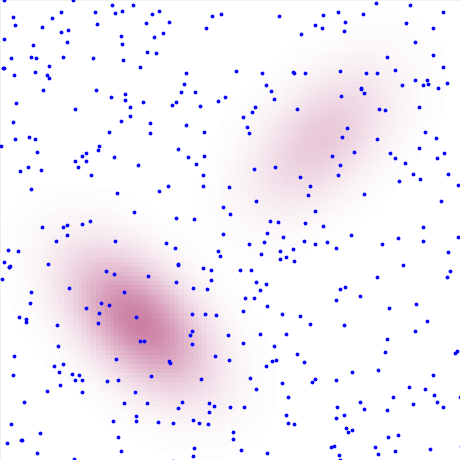}
        \includegraphics[width=0.15\columnwidth]{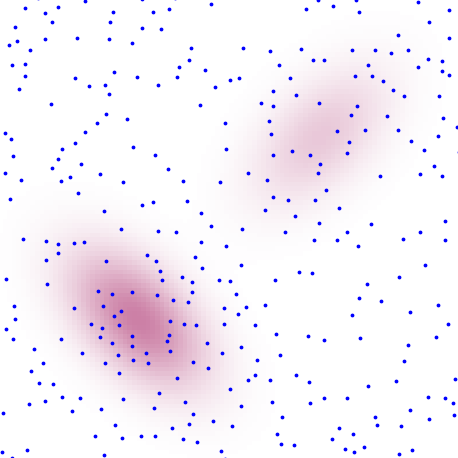}
        \includegraphics[width=0.15\columnwidth]{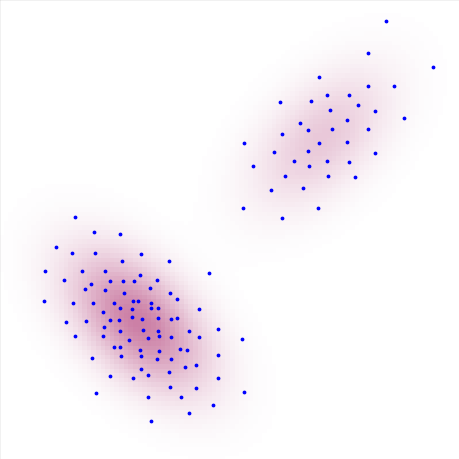}
        \includegraphics[width=0.15\columnwidth]{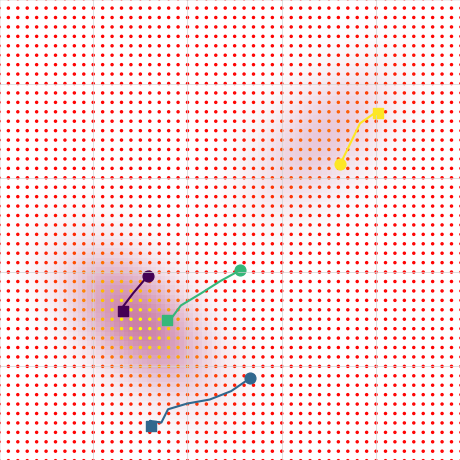}
    \end{minipage}
    
    \caption{Two toy 2D Gaussian densities. Upper: unimodal, lower: bimodal. Left to right: iteration 0, 5, 60 and selected particle trajectories (dot: start, square: end). Pink contours are ground truth density; blue dots denote negative charges, red dots denote positive charges. See Appendix.\ref{app:toy_examples} for details.}
    \label{fig:toy_Gaussians}
\end{figure}

\paragraph{Gaussian densities} 
Two examples, a unimodal Gaussian and a bimodal Gaussian mixture (mixing proportions 0.7/0.3), are presented in Fig.\ref{fig:toy_Gaussians}. Fig.\ref{fig:toy_Gaussians}(a) starts from a uniform prior which partially overlaps with the target, converges fast and the inferred shape (third from left) aligns well with ground truth (pink contours). Fig.\ref{fig:toy_Gaussians}(b) identifies the two modes and their mixing proportions very well. Unlike many MCMC algorithms which encourage particles to move from low to high density regions (either following the steepest gradient direction or a probabilistic MH step), we observe that, the interacting particles following the deterministic EParVI algorithm don't restrict themselves to follow the gradient.

\begin{figure}[ht]
    \centering
    \begin{minipage}[b]{1.0\columnwidth}
        \centering
        \includegraphics[width=0.15\columnwidth]{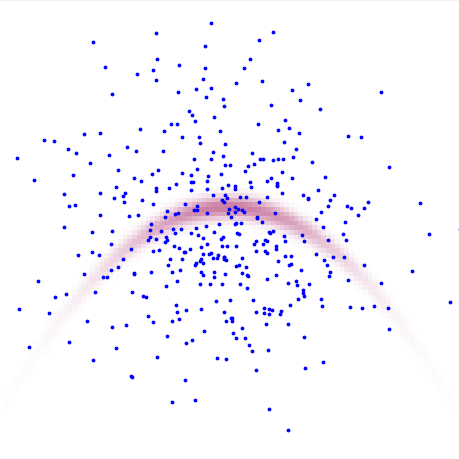}
        \includegraphics[width=0.15\columnwidth]{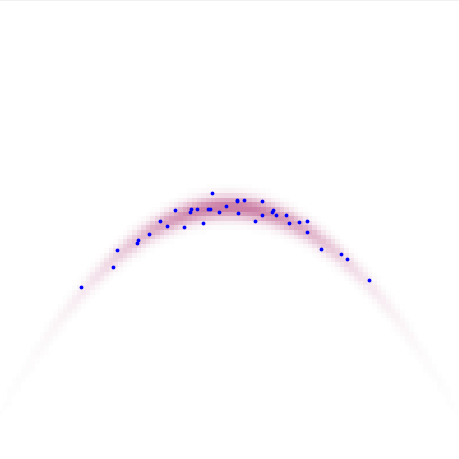}
        \includegraphics[width=0.15\columnwidth]{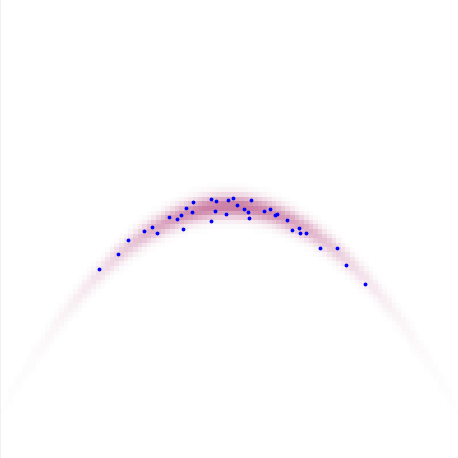}
    \end{minipage}
    
    \begin{minipage}[b]{1.0\columnwidth}
        \centering
        \includegraphics[width=0.15\columnwidth]{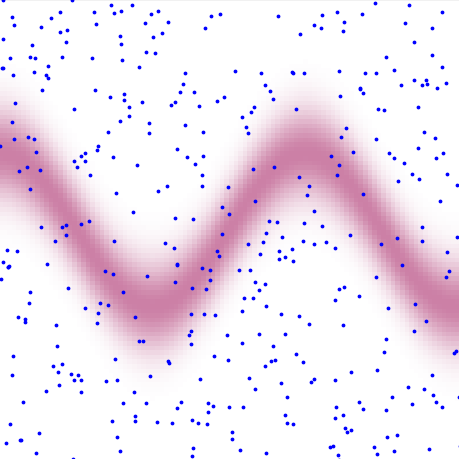}
        \includegraphics[width=0.15\columnwidth]{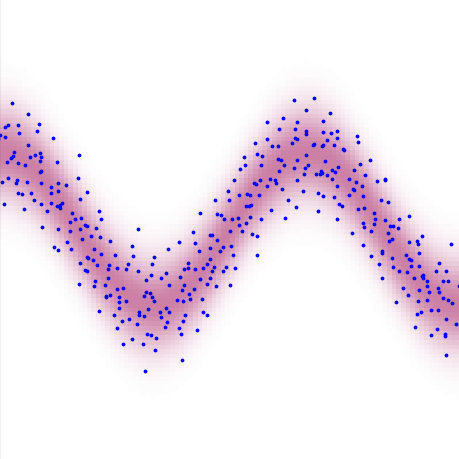}
        \includegraphics[width=0.15\columnwidth]{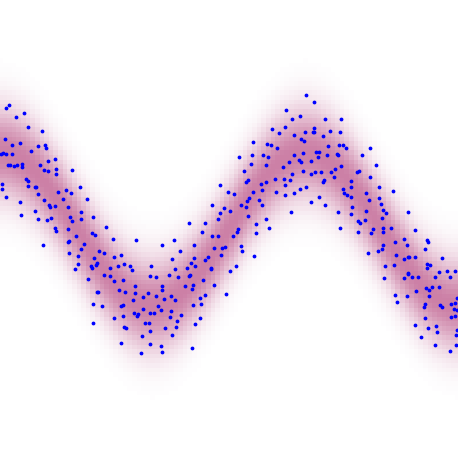}
    \end{minipage}

    \begin{minipage}[b]{1.0\columnwidth}
        \centering
        \includegraphics[width=0.15\columnwidth]{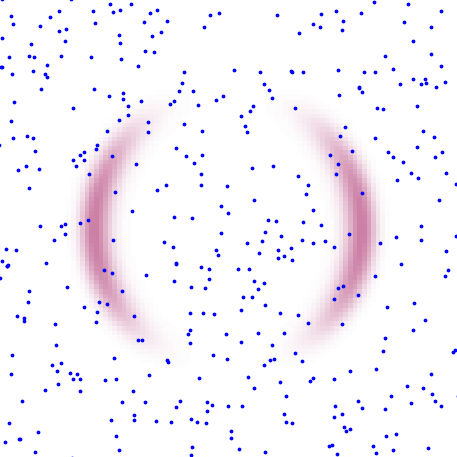}
        \includegraphics[width=0.15\columnwidth]{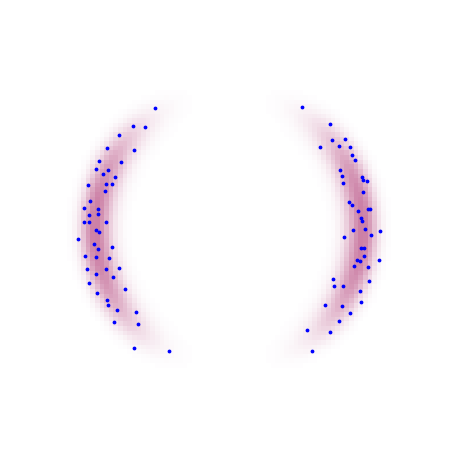}
        \includegraphics[width=0.15\columnwidth]{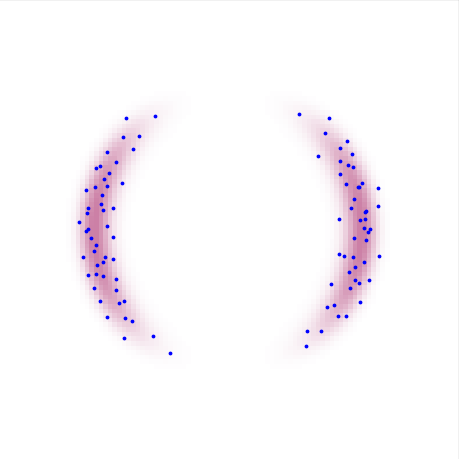}
    \end{minipage}
    
    \caption{Three other toy 2D densities. Left to right: iteration 0, 40, 60. See Appendix.\ref{app:toy_examples} for details.}
    \label{fig:other_densities}
\end{figure}

\paragraph{Other densities} We also test the EParVI algorithm on three other toy densities, i.e. a moon-shaped density, double-banana, a wave density, which are commonly used in other VI benchmark tests \cite{haario1999adaptive,chen2018unified,liu2019understanding,Wang2021EVI}. The results are shown in Fig.\ref{fig:other_densities}. With different initialisation, the EParVI algorithm is able to transport particles to the target regions while maintaining diversity. Dispersion of particles is secured by the repulsive forces, which helps prevent model collapse.

\subsection{Neal's funnel} Neal's funnel distribution, as written in Eq.\ref{eq:neals_funnel2}, has been challenging to infer due to its highly anisotropic shape and exponentially increasing variance - the funnel’s neck is particularly sharp, making it a benchmark for testing many MCMC and VI methods\cite{Neal2003slice}. 

\begin{equation} \label{eq:neals_funnel1}
p(x_1, x_2) = p(x_1|x_2) p(x_2)
\end{equation}

\noindent where $
p(x_2) = \mathcal{N}(x_2|0, \sigma^2), p(x_1|x_2) = \mathcal{N}(x_1|0, \exp(x_2/2))
$. The joint pdf is therefore:

\begin{equation} \label{eq:neals_funnel2} \tag{\ref{eq:neals_funnel1}b}
p(v, x) = \frac{1}{\sqrt{2 \pi \cdot \sigma^2}} \exp \left( -\frac{x_2^2}{2 \cdot \sigma^2} \right) \cdot \frac{1}{\sqrt{2 \pi \exp(x_2/2)}} \exp \left( -\frac{x_1^2}{2 \exp(x_2/2)} \right)
\end{equation}

We test 6 methods, i.e. MH, HMC, LMC, SVGD EVI and EParVI, on the 2D differentiable geometry with $\sigma=3$, compare their inference quality, as measured by two metrics, i.e. the squared maximum mean discrepancy $MMD^2$ \cite{Arbel2019,Wang2021EVI} and the average negative log-likelihood (\textit{avg NLL}), as well as their runtimes in each iteration. While MH and HMC produce samples sequentially, Langevin, SVGD, EVI and EParVI generate samples iteratively and are started with the same initialisation.

\begin{figure}[ht]
    \centering
    \begin{minipage}[b]{1.0\columnwidth}
        \centering
        \includegraphics[width=0.19\columnwidth]{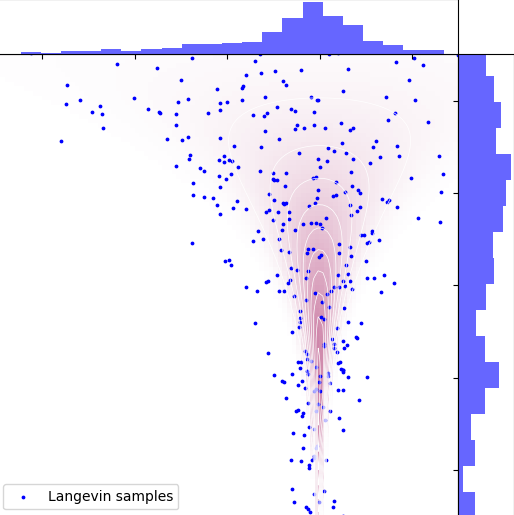}
        \includegraphics[width=0.19\columnwidth]{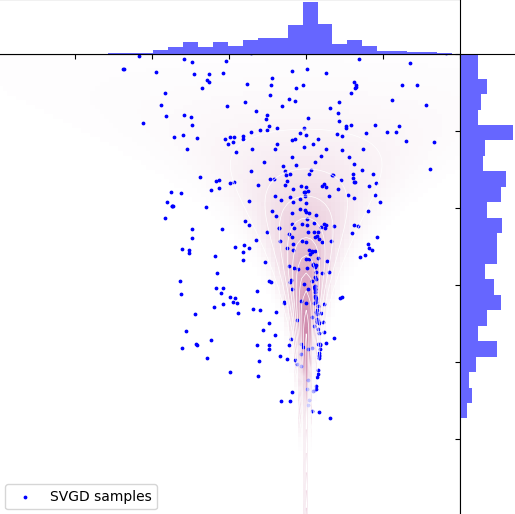}
        \includegraphics[width=0.19\columnwidth]{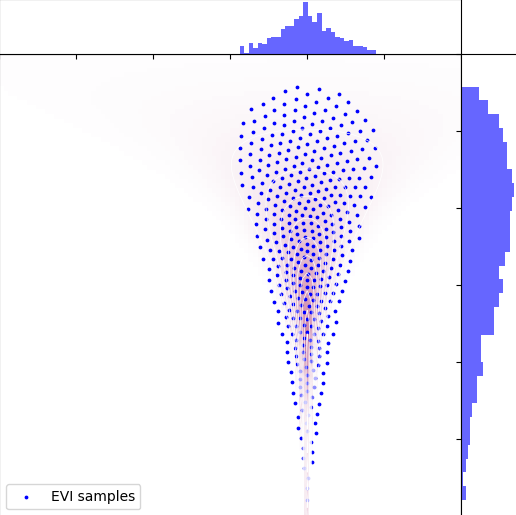}
        \includegraphics[width=0.19\columnwidth]{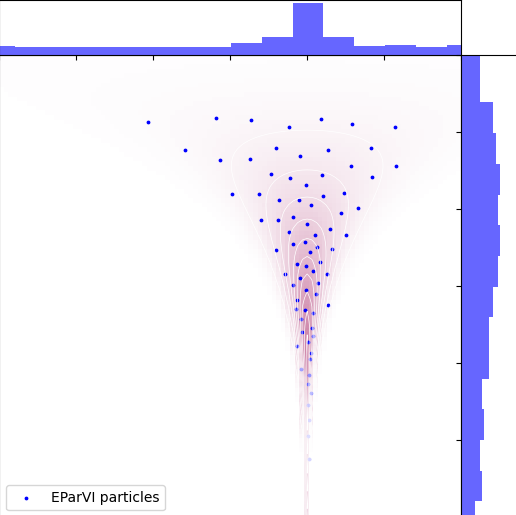}
        \includegraphics[width=0.19\columnwidth]{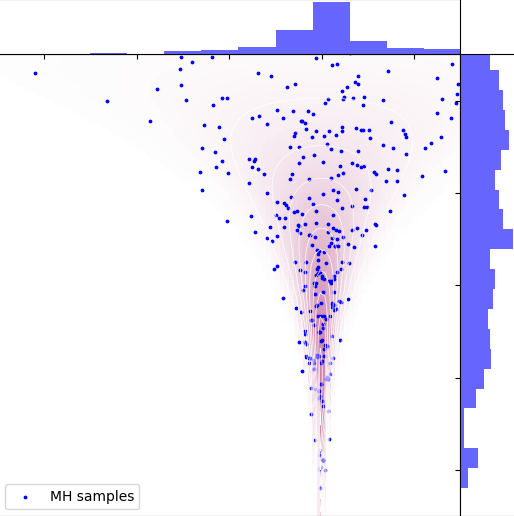}
    \end{minipage}

    \begin{minipage}[b]{1.0\columnwidth}
        \centering
        \includegraphics[width=0.19\columnwidth]{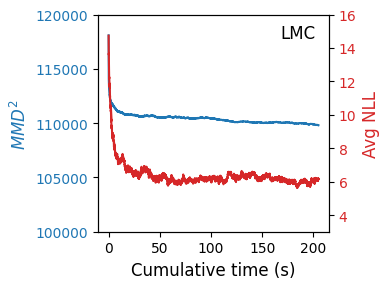}
        \includegraphics[width=0.19\columnwidth]{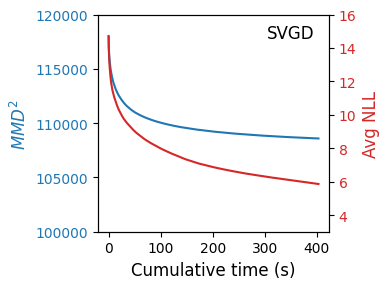}
        \includegraphics[width=0.19\columnwidth]{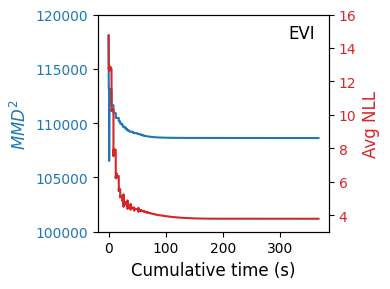}
        \includegraphics[width=0.19\columnwidth]{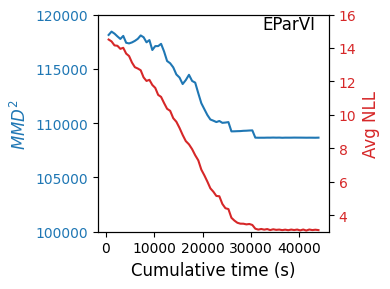}
        \includegraphics[width=0.19\columnwidth]{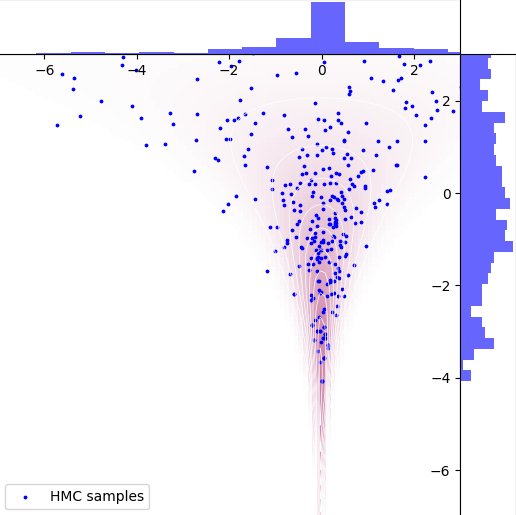}
    \end{minipage}
    
    \caption{Inference of the Neal's funnel using six methods. Upper left to right: LMC, SVGD, EVI, EParVI, and MH samples. Lower left to right: corresponding $MMD^2$, $NLL$ values over runtime and HMC samples. MH and HMC take seconds to run. See Appendix.\ref{app:neals_funnel_implementation_details} for details.}
    \label{fig:neals_funnel}
\end{figure}

Results presented in Fig.\ref{fig:neals_funnel} show that, with a medium number of particles (i.e. 400), MH and LMC are the fastest in terms of speed, SVGD and EParVI are among the slowest due to large number of iterations between particles. EVI is fastest among these particle methods. HMC, being fast and accurate, is considered as the golden method for sampling complex geometries\cite{Hoffman2014}; we therefore use the HMC samples as a reference for evaluating $MMD^2$. Key observations are: while all samples show some similarity to the HMC samples, EParVI samples notably yield the smallest average NLL, which implies they are better quasi samples following the target distribution, as also evidenced by the upper row figures.

\subsection{Bayesian Logistic Regression} 

We also apply the proposed EParVI, along with other 4 counterparts, i.e. logistic regression via maximum likelihood (MLE-LR), HMC, SVGD and EVI, to infer the coefficients of a low-dimensional Bayesian logistic regression (BLR) model, using the Iris dataset $\{(\textbf{x}_i, y_i)\}_{i=1}^N$ which has $N=150$ observations with binary labels $y_i$, and each instance $\textbf{x}_i$ consists of 4 features. The logistic regression model is $p(y_i = 1|\textbf{x}_i, \boldsymbol{\omega}) = \frac{1}{1 + \exp(-\boldsymbol{\omega}^T \textbf{x}_i)}$. We use standard Gaussian prior for regression coefficients, i.e. $\boldsymbol{\omega} \sim \mathcal{N}(\boldsymbol{0}, \boldsymbol{I})$. The posterior distribution of coefficients $\boldsymbol{\omega}$ is then (details please refer to Appendix.\ref{app:BLR}):

\begin{equation} \label{eq:BLR_target}
p(\boldsymbol{\omega}|\boldsymbol{y}, \textbf{X}) \propto \prod_{i=1}^N \left[ \frac{1}{1 + \exp(-\boldsymbol{\omega}^T \textbf{x}_i)} \right]^{y_i} \left[ 1 - \frac{1}{1 + \exp(-\boldsymbol{\omega}^T \textbf{x}_i)} \right]^{1 - y_i} \exp \left( -\frac{1}{2\alpha} \boldsymbol{\omega}^T \boldsymbol{\omega} \right)
\end{equation}

\noindent which is used as the target for our sampling methods.

\begin{figure}[!ht]
    \centering
    \begin{minipage}[b]{1.0\columnwidth}
        \centering
        \includegraphics[width=0.30\columnwidth]{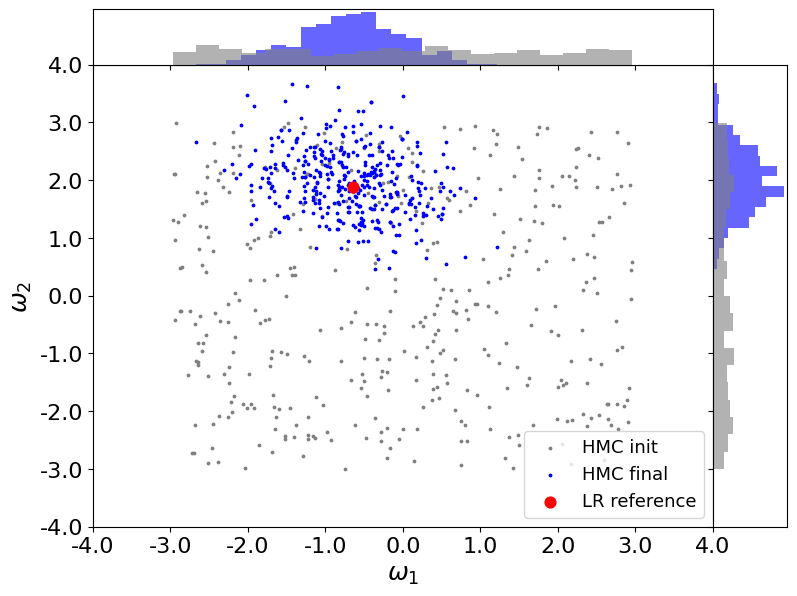}
        \includegraphics[width=0.30\columnwidth]{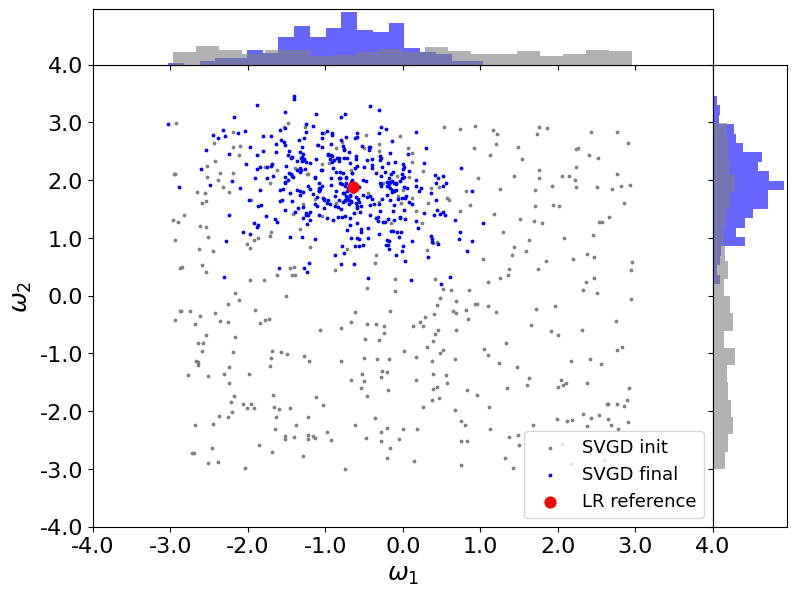}
    \end{minipage}

    \begin{minipage}[b]{1.0\columnwidth}
        \centering
        \includegraphics[width=0.30\columnwidth]{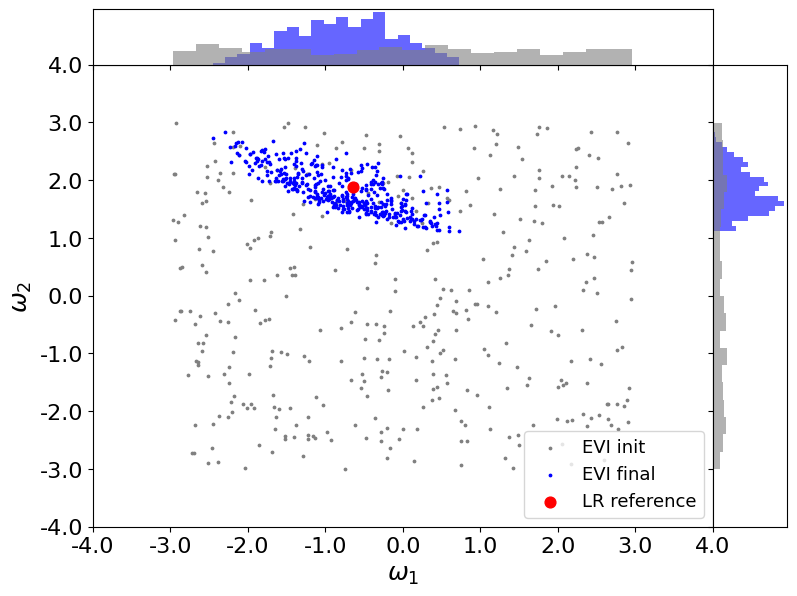}
        \includegraphics[width=0.30\columnwidth]{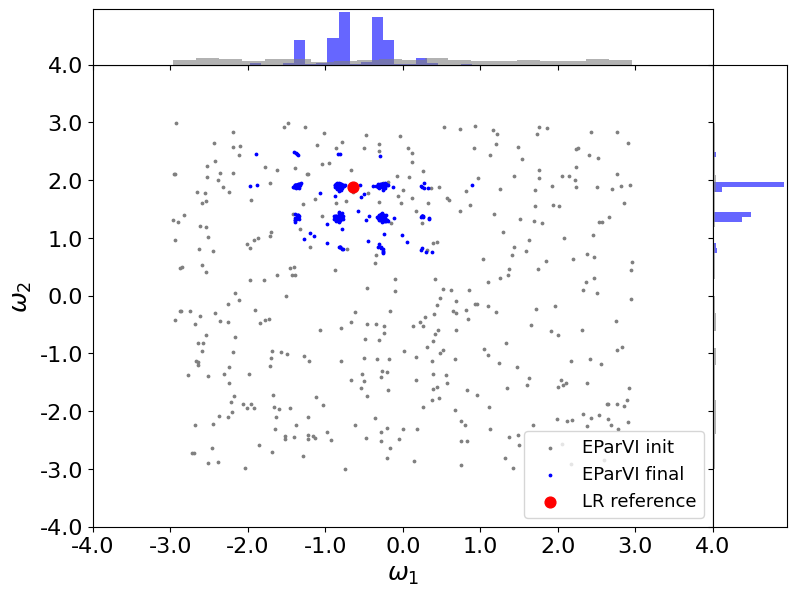}
    \end{minipage}
    
    \caption{Inference of BLR model. Upper left to right: HMC and SVGD samples. Lower left to right: EVI and EParVI samples. MH and HMC take seconds to run. See Appendix.\ref{app:BLR} for details.}
    \label{fig:BLR_samples}
\end{figure}

The initial and final samples are visualised for the two dimensions ($\omega_1$,$\omega_2$) in Fig.\ref{fig:BLR_samples}. All 4 methods yield reasonably well inference quality - all terminal particles cluster around the ground truth point with some dispersion representing uncertainties. The EParVI particles, in particular, have reduced uncertainty as compared to other methods. A comparison of the statistics are made in Table.\ref{tab:BLR_comparison}, in which we observe HMC yields results closest to the MLE-LR estimates. Particle methods give less accurate estimates yet fast speeds such as SVGD and EVI. EParVI achieves smallest NLL but less impressive mean estimates; it also differs from the HMC samples due to less iterations and the coarse granularity used in meshing.

\begin{table}[ht]
    \centering
    \renewcommand{\arraystretch}{1.2}  
    \begin{small}  
    \begin{tabular}{|c|c|c|c|c|c|c|}
    \hline
    \textbf{Method} & \(\hat{\omega}_1\) & \(\hat{\omega}_2\) & \(\hat{\omega}_3\) & \(\hat{\omega}_4\) & \textbf{$NLL$/$MMD^2$} & runtime \\
    \hline
    MLE-LR & -0.64  &1.89  &-1.75 &-1.67  &- & $<1$ second \\
    \hline
    HMC &-0.69 & 1.98  &-1.84 &-1.86  &-  & $<30$ seconds\\
    \hline
    SVGD &-0.76  & 1.87  & -1.74  & -1.67 & 12.83 / 2.41  & $<100$ seconds\\
    \hline
    EVI & -0.78  & 1.81  & -1.63  & -1.56 & 11.39 / 10.14 & $<100$ seconds \\
    \hline
    EParVI &-0.64 &1.62  &-1.50   &-1.42  & 11.32 / 23.25 & $\sim 10^4$ seconds \\
    \hline
    \end{tabular}
    \end{small}
    \caption{Comparison of inference methods for the BLR problem. Sample mean estimates of coefficients are presented for all but MLE-LR. With these inferred coefficients, the test accuracy for all cases is $100\%$. The runtime is estimated on \textit{Colab TPU v2}. NB: the EParVI runtime can be massively improved by caching the magnitudes of positive charges for all time steps.}
    \label{tab:BLR_comparison}
\end{table}

\subsection{Inference of the Lotka-Volterra dynamical system} \label{sec:LV_systm}

\begin{figure}[ht]
    \centering
    \includegraphics[width=0.40\columnwidth]{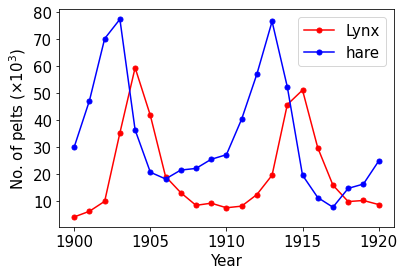}
    \caption{Temporal evolution of the LV system.}
    \label{fig:LV_data}
\end{figure}

Here we look at the Lotka-Volterra (LV) ecological system whose dynamics can be represented by the following two ODEs:

\begin{equation} \label{Eq:LV_dynamics}
\begin{aligned}
    \frac{dx}{dt} &= ax - bxy \\
    \frac{dy}{dt} &= cxy - dy \\
\end{aligned}
\end{equation}

\noindent where $x$ and $y$ are the numbers of hare and lynx pelts. We are interested in inferring the for inferring the 4-dimensional parameter $\theta=(a,b,c,d)$ based on 21 observed real-world data as plotted in Fig.\ref{fig:LV_data}. This can be formulated as a Bayesian inference problem with the posterior:

\begin{equation} \label{Eq:LV_posterior}
    p(\theta|X',Y') = p(a,b,c,d|X',Y') = \frac{p(X',Y'|a,b,c,d) p(a)p(b)p(c)p(d)}{Z}
\end{equation}

\noindent where $Z$ is the normalising constant. The observed trajectories $(X',Y')=\{(x'_i,y'_i)\}_{i=1}^{21}$, and independence of parameters is assumed. The priors are specified as:

\begin{equation} \label{Eq:LV_prior}
a, d \sim U(0.001, 1.0), \hspace{0.25cm} b, c \sim U(0.001, 0.05)
\end{equation}

\noindent and likelihood:

\begin{equation} \label{eq:LV_likelihood}
    \log x'(t_i) \sim \mathcal{N}(\log x(t_i), 0.25^2), \hspace{0.25cm} \log y'(t_i) \sim \mathcal{N}(\log y(t_i), 0.25^2)
\end{equation}

\noindent where we have for simplicity assumed known \footnote{These known values are referenced from \cite{LV_Carpenter}}, homogeneous state noise level $\sigma=0.25$, and known initial conditions $x_0=33.956, y_0=5.933$. Assuming independence between the states writes the likelihood as:

\begin{multline} \label{Eq:LV_likelihood_func}
    p(X',Y'|a,b,c,d) = p(X'|a,b,c,d)p(Y'|a,b,c,d)
    = \prod_{i=1}^N p(x_i'|a,b,c,d) \prod_{i=1}^N p(y_i'|a,b,c,d) \\
    = \prod_{i=1}^N \frac{1}{\sqrt{2\pi}\sigma} \frac{1}{x_i'} exp[-\frac{(\log x_i' - \log x_i)^2}{2\sigma^2}] \frac{1}{\sqrt{2\pi}\sigma} \frac{1}{y_i'} exp[-\frac{(\log y_i' - \log y_i)^2}{2\sigma^2}] 
\end{multline}

\noindent where $x(t_i)$ and $y(t_i)$ are the solutions to Eq.\ref{Eq:LV_dynamics},  $x'(t_i)$ and $y'(t_i)$ are the observed data in Fig.\ref{fig:LV_data}. 

Substituting the priors (Eq.\ref{Eq:LV_prior}) and likelihood (Eq.\ref{Eq:LV_likelihood_func}) into the posterior (Eq.\ref{Eq:LV_posterior}), we obtain the probabilistic model $p(\theta|X',Y')$, which is the target of our inference practice. To make the learning more numerically stable \footnote{The chain product in the posterior Eq.\ref{Eq:LV_posterior} leads to small absolute values of the target, and thus requires large positive magnitude $q$.}, we set the positive charge magnitudes \footnote{Note that, it is crucial for the charge magnitudes $q_i$, where $i \in \{1,2,...M^{neg}\}$, and $q_{i'}$, where $i' \in \{1,2,...M^{pos}\}$, to remain positive. While we use $q_i$=1.0 in all experiments; the magnitudes of positive charges $q_{i'}$, however, can be negative in scenarios e.g. $q_{i'}=\log p(\textbf{x}_{i'})$. Therefore, an offset is made, i.e. $q_{i'}=\log p(\textbf{x}_{i'})-min[\log p(\textbf{x}_{i'})]$ to ensure positivity. Further, at grid points which results in negative values of ($x(t),y(t)$), we set $q_{i'}$=0.} $q_j=q \log p(\theta_j|X',Y')$ where $q$ is a scaling factor which specifies the maximum magnitude, as shown in Eq.\ref{Eq:LV_log_posterior} (Appendix.\ref{app:LV_system}). 

\begin{figure}[ht] 
    \centering
    \begin{subfigure}[b]{0.50\columnwidth}
        \includegraphics[width=\linewidth]{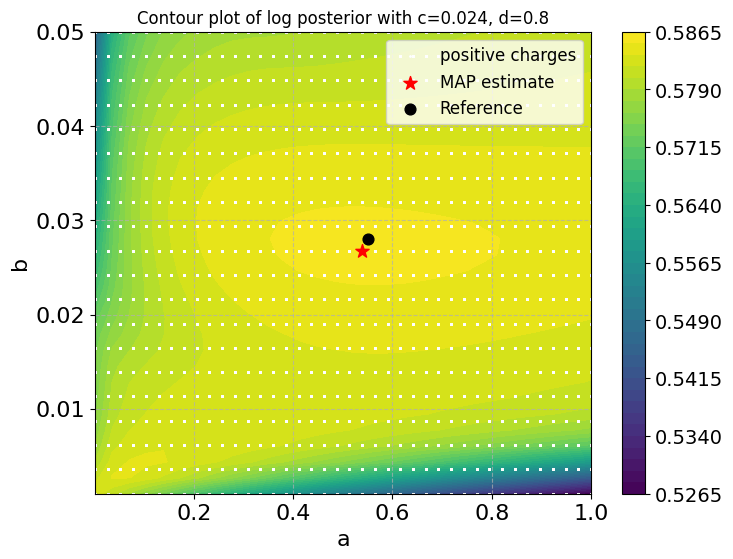}
    \end{subfigure}
    \hfill 
    \begin{subfigure}[b]{0.48\columnwidth}
        \includegraphics[width=\linewidth]{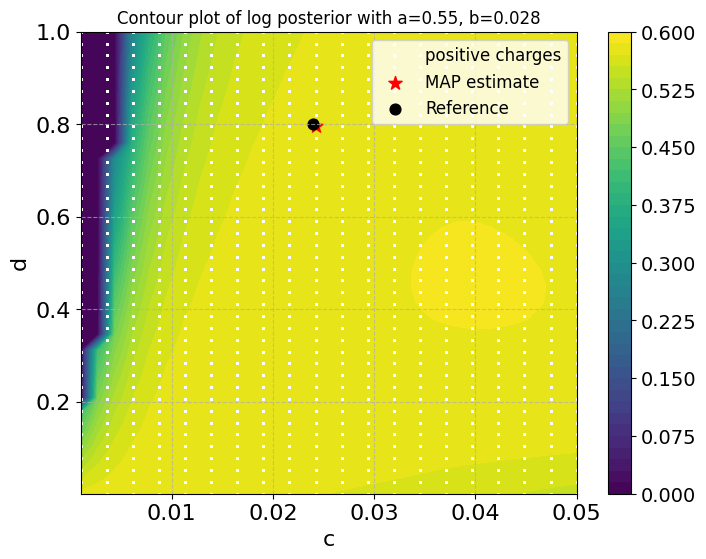}
    \end{subfigure}
    \caption{Log posterior contour plots. Left: snapshot at $c=0.024,d=0.8$; right: snapshot at $a=0.55,b=0.028$.}
    \label{fig:LV_log_posterior}
\end{figure}

Before applying EParVI, we do a quick check of the probabilistic model. We evaluate $p(\theta|X',Y')$ over all mesh grid points, and compare its maximum, i.e. the maximum a posterior (MAP) estimate, with reference values. The MAP estimate is
\[
\hat{\theta}^{MAP} = argmax_j p(\theta_j|X',Y') \equiv argmax_j \log p(\theta_j|X',Y') = [\overset{a}{0.539}, \overset{b}{0.027}, \overset{c}{0.024}, \overset{d}{0.795}]
\]
\noindent which is very close to the ground truth value from literature \cite{LV_Carpenter}. This confirms that, with the current meshing granularity, the peak of the positive charge magnitude aligns with ground truth. The whole approximate landscape of $p(\theta|X',Y')$ is visualised in Fig.\ref{fig:LV_log_posterior}, and the marginal distributions in Fig.\ref{fig:LV_marginal_log_posterior}. We observe that, the log posterior surface is relatively flat, which could potentially impact the identifiability of modes.

\begin{figure}[ht]
    \centering
    \includegraphics[width=1.0\columnwidth]{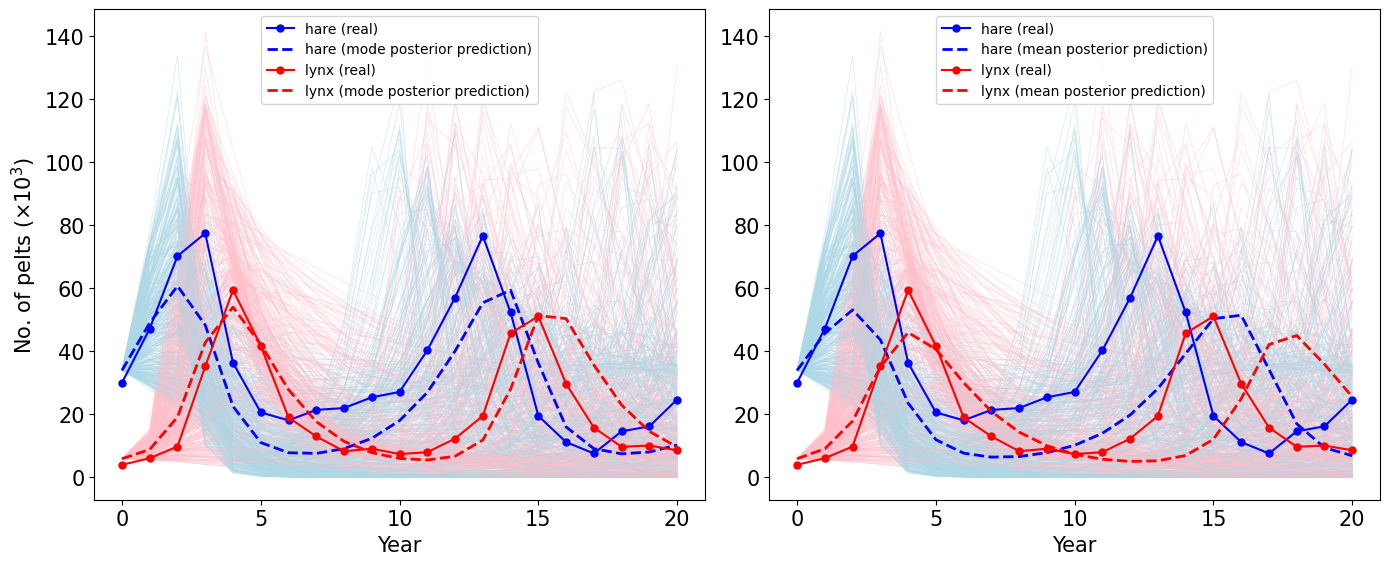}
    \caption{EParVI posterior-based trajectory predictions. Left: predictions made using mean posterior, right: predictions made using posterior mode. In both figures, light blue and pink lines represent the predictions for hare and lynx, respectively, made by all final particles.}
    \label{fig:LV_posterior_predictive_EParVI}
\end{figure}

Then we run 5 methods, i.e. MAP, HMC, LMC, SVGD and EParVI, to infer the posterior $p(\theta|X',Y')$, using similar model settings \footnote{The same priors are used in all six methods; HMC, however, for convenience of implementation (mainly due to the negativity of some trajectory values at impossible $\theta$ locations, which results in error when taking log), a different likelihood is used: $x'(t_i) \sim \mathcal{N}(x(t_i),\sigma^2)$ and $y'(t_i) \sim \mathcal{N}(y(t_i),\sigma^2)$.}, and compare their performances. For EParVI, we first simulate the LV system at a grid of mesh points (Fig.\ref{fig:LV_log_posterior}), obtaining $(X,Y)$ and substituting $(X,Y)$ and $(X',Y')$ into the log posterior to obtain the positive charge magnitudes at these mesh points. These magnitudes are cached for later query. Then we randomly initialise $\theta=(a,b,c,d)$, i.e. the coordinates of negative charges, and calculate the pairwise distances between each negative charge and all other charges, further the repulsive and attracting forces (by querying the positive charge magnitudes) acting on each negative charge. The overall force is then substituted into the updating rule to give the displacement for each negative charge. We run the EParVI algorithm till stability using 400 negative and 640000 positive particles, the initial and final configurations of the particles, along with their empirical marginal distributions, are presented in Fig.\ref{fig:LV_prior_and_posterior_EParVI} (more details see Fig.\ref{fig:LV_empirical_marginal_dist} in Appendix.\ref{app:LV_system}). It is seen that, the inferred geometries for $b,c$ are narrow and peaky, while those for $a,d$ are flat and widely distributed; final particles sit in the high probability regions of the posterior, which is consistent with the posterior snapshots in Fig.\ref{fig:LV_log_posterior} - particles along the dimensions $b$ and $c$ are moer concentrated than in dimensions $a$ and $d$. The relatively wide spread of the posterior distributions of $a$ and $d$ may be due to the fact that \cite{LV_Kaji}, realizations of the LV system are more sensitive to the interaction parameters $b$ and $c$, while less sensitive to $a$ and $b$. The posterior-based predictions are presented in Fig.\ref{fig:LV_posterior_predictive_EParVI}, from which we observe loosely good match between the empirical posterior mode-based predictions and real trajectories, with some lifts and shifts. The predictions made using all final particles, which represent uncertainties, exhibit large variations.

\begin{figure}[ht] 
    \centering
    \begin{subfigure}[b]{0.50\columnwidth}
        \includegraphics[width=\linewidth]{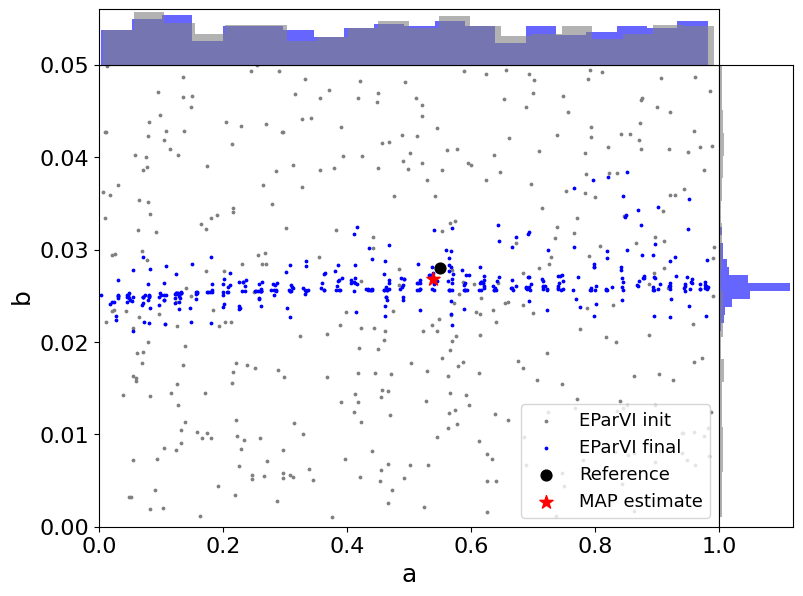}
    \end{subfigure}
    \hfill 
    \begin{subfigure}[b]{0.48\columnwidth}
        \includegraphics[width=\linewidth]{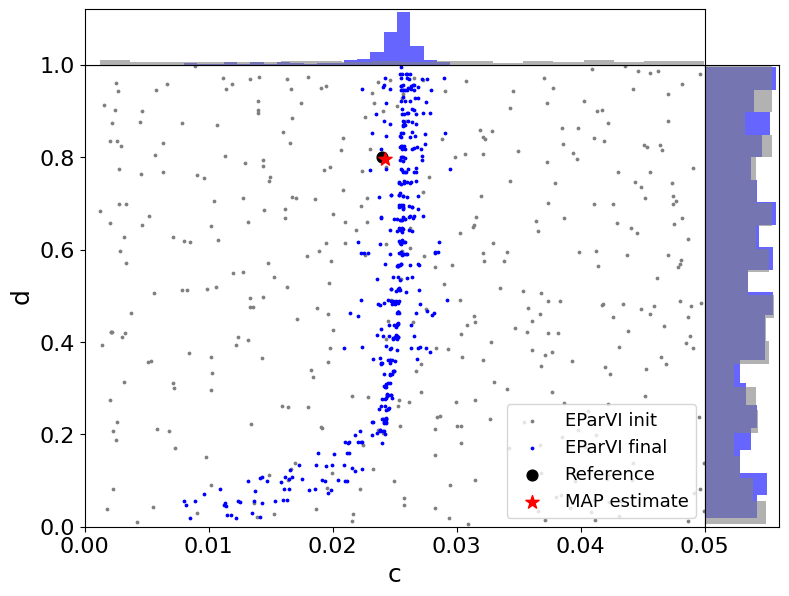}
    \end{subfigure}
    \caption{EPaVI: initial and final particle distributions.}
    \label{fig:LV_prior_and_posterior_EParVI}
\end{figure}

A comparison of the inference results is made in Table.\ref{tab:LV_inference}. We observe that, the EParVI estimations are close to the reference values, particularly in dimensions $b$ and $c$. It, however, also occupies the longest runtime. EParVI outperforms SVGD and LMC; the optimal choice, balancing accuracy and efficiency, however is HMC.

\begin{table}[h!]
\begin{center}
\small
\begin{threeparttable}
\begin{tabular}{m{3.1cm} m{1.4cm} m{1.4cm} m{1.4cm} m{1.4cm} m{1.4cm}} 
  \toprule 
  Method\textbackslash Quantity & $a$ & $b$ & $c$ & $d$ & runtime(s)\tnote{1} \\ 
  \midrule
  EParVI (mean/mode) & 0.487/\textbf{0.549} & 0.026/0.026 & \textbf{0.024}/0.025 & 0.529/0.645 & 221400 \\ 
  SVGD (mean/mode) & 1.017/1.029 & 0.076/0.073 & 0.045/0.032 & 1.025/0.978 & 8662 \\   
  LMC (mean/mode) &0.482/0.556 &0.028/0.019 &0.021/0.021 &0.555/0.651 &24178 \\ 
  HMC (mean/mode) & 0.566/0.570 &\textbf{0.028}/0.027 &0.023/0.023 &\textbf{0.763}/0.759 &\textbf{366} \\ 
  MAP (EParVI mesh) & 0.539 & 0.027 & 0.024 & 0.795 & - \\ 
  Reference\tnote{2} & \underline{0.55} & \underline{0.028} & \underline{0.024} & \underline{0.80} & - \\ 
  \bottomrule
\end{tabular}
\begin{tablenotes}
  \item[1] These methods can differ in numbers of samples and iterations, see Appendix.\ref{app:LV_system}.
\end{tablenotes}
\caption{Inference results of the LV system coefficients $\theta=[a,b,c,d]$ using six methods.}
\label{tab:LV_inference}
\end{threeparttable}
\end{center}
\end{table}

\section{Discussions} \label{sec:discussions}
EParVI is built on the principles that charges repel and attract each other, in regions of uniform density, charges are distributed uniformly; in regions where density varies, the clustering of negative charges follows the density distribution. The repulsive forces between negative charges ensure particle diversity and avoid mode collapse. These principles ensure the correctness and diversity of the inferred geometry. From an optimisation point of view, each update reduces global force imbalance by repositioning the particles in accordance with the force field \cite{Schmaltz2010}; from an energy perspective, we are implicitly minimizing the overall electric potential which serves as the variational objective. Energy is minimized when forces are in equilibrium (i.e. system is electrically neutral)\cite{Teuber_energy}. More detailed discussions are delegated to Appendix.\ref{app:other_discussions}.

\paragraph{Computation in high-dimensions} EParVI suffers from curse of dimensionality. The quality of sampling and inference results relies on the granularity of meshing. Finer mesh grid is needed for complex geometries (e.g. with large curvature). However, the computational complexity increases exponentially with dimension $m$. Efforts can be saved by, e.g. pre-calculating the probability matrix at grid points, as well as pairwise forces between cells, and storing them for later query. Fast Fourier transform techniques \cite{Gwosdek2014} have also been developed. It would be beneficial to leverage the fast inference capability of a trained neural network to approximate the force field as well.

\paragraph{Local optima} This can lead to ambiguous distribution of the negative charges; adding perturbations to the particle in each iteration may help, e.g. projecting the particle onto the closest horizontal or vertical line connecting the grid points \cite{Schmaltz2010}. We propose injecting some random noise to the move proposal, e.g. every $k$ steps, to avoid getting stuck in local optima:

\begin{equation} \label{eq:loca_optima}
    \textbf{x}^{t+1}_j = \textbf{x}^{t}_j + \tau' \Delta \textbf{x}^t_j + \epsilon^t, \hspace{0.1cm} \text{every $k$ steps}
\end{equation}

\noindent where $\epsilon_j^t \sim \mathcal{N}(0,\sigma^2)$. This is equivalent to saying $\textbf{x}^{t+1}_j \sim \mathcal{N}(\textbf{x}^{t}_j + \tau' \Delta \textbf{x}^t_j, \epsilon_j^t)$. The proposed move $\tau' \Delta \textbf{x}^t_j=\tau \textbf{F}^t_j$ for the Euler scheme, and $\Delta \textbf{x}^t_j=\textbf{F}_j \Delta t^2 + \Delta \textbf{x}^{t-1}_j$ for the Verlet and damped Verlet schemes. 

\paragraph{Probabilistic move} If, in an iteration, the particle takes a large step and jumps to a low density region, it may get trapped in that region due to small drag force (e.g. the extra regularisation forces outweighing the attracting force \cite{Schmaltz2010}). We thus suggest taking a probabilistic move to avoid occasionally large move, and assess the quality of each move via an MH step to decide if this move is to be executed:

\begin{equation} \label{Eq:SVGD_update_rule_MH}
 \textbf{x}_j^{t+1} =
    \begin{cases}
         \textbf{x}^{t}_j + \tau \Delta \textbf{x}^t_j, & \text{with probability } p_{accept} \\
        \textbf{x}_j^t, & \text{else }
    \end{cases}
\end{equation}

\noindent where 

\begin{equation} \label{Eq:SVGD_Metropolis_accept_prob}
    p_{accept} = min\{1, \frac{p(\textbf{x}^{t}_j + \tau \Delta \textbf{x}^t_j)}{p(\textbf{x}^{t}_j)}\} \\
\end{equation}

\noindent in which the ratio doesn't require a fully know density. Using this acceptance-rejection criteria, we can then decide if a move proposal $\tau \Delta \textbf{x}^t_j$ should be accepted or not. The use of an MH step, however, is two-side: while it prevents the particle of consideration from taking dangerous move in the current iteration, it may break the strategy for promoting global force equilibrium. It also induces extra cost.

\paragraph{EParVI diagnostics} First, we need to ensure positivity of the charge magnitudes $q_i$, where $i \in \{1,2,...M^{neg}\}$, and $q_{i'}$, where $i' \in \{1,2,...M^{pos}\}$. Negative magnitudes can result, e.g. in scenarios where $q_{i'}=\log p(\textbf{x}_{i'})$ (e.g. the LV dynamical system inference problem). Therefore, adjustment (offset) is needed. Further, there may be other constraints imposed by the problem itself, e.g. positive trajectories in the LV dynamical system inference problem, which requires special treatment (e.g. $q_{i'}$=0) of the positive charge magnitudes at grid points yielding negative ($x(t),y(t)$) values. In each step, we monitor the number of free-moving negative charges which are sitting within the initialisation regime; large number of negative charges falling out of this region may indicate potentially wrong specification of the probabilistic model, or erroneous calculation of the positive charge magnitudes (e.g. negative values appear in magnitudes), or simply the repulsive forces is way bigger than the attracting force (i.e. too small positive charge magnitudes). Correctness of the probabilistic model can be checked via other sampling methods such as MCMC, or by comparing the MAP estimate with point ground truth values (if ground truth values are available). The imbalance between repulsive and attracting forces can be checked by running the simulation for several trial steps, monitoring both values, and increasing/decreasing the charge magnitudes accordingly.

\section{Conclusion and future work} \label{sec:conclusion_and_future}

\paragraph{Conclusion} We propose a variational sampling and inference method, EParVI, which adheres to physical laws. EParVI simulates an interacting particle system in which positive particles are fixed at grid points and negative particles are free to move. With positive charge magnitudes proportional to the target distribution, negative particles evolve towards a steady-state approximating the target distribution; particles repositioning follows a discretised Newton's second law, and the force acts in accordance with the electric fields described by the Poisson's equation. It features physics-based, gradient-free, deterministic dynamics sampling, with traits of simplicity, non-parametric flexibility, expressiveness and high approximation accuracy; as a mesh-based method, it also suffers from curse of dimensionality. 

Experimental results confirm the effectiveness of EParVI. It exhibits superior performance in inferring low-dimensional, complex densities; it also achieves comparable performance as other MCMC and ParVI methods in real-world inference tasks such as Bayesian logistic regression and dynamical system identification. This is the first time an inference problem is fully formulated as a physical simulation process.

\paragraph{Future work} There are theoretical aspects which can be explored. For example, the convergence of the IPS towards equilibrium with theoretical guarantees. This can be routed to using the variational potential energy objective. Also, improving the computational efficiency is crucial for its applicability. Further, this technique can be potentially used for optimisation. However, sampling differs from optimisation in that, we can query the target function (i.e. the density, possibly partially known) as many times as needed; in optimisation, however, one typically avoids to perform functional evaluations too many times as it's expensive.

\paragraph{Code availability} Full codes are available at: \url{https://github.com/YongchaoHuang/EParVI}

\printbibliography

\appendix

\section{A hypothetical treatment of Coulomb's law in \textit{d}-dimension} \label{app:coulbomb_law_nd}

Here we aim to derive the forces between two charges in low and high dimensions. We first present the well-known force formula in 3D (Section.\ref{app:force_3D}), then move to general $\textit{d}$ dimensions (Section.\ref{app:force_ND}), solve the 3D Poisson's equation (Section.\ref{app:Poissons_equation_3D}), and solve the Poisson's equation in $\textit{d}$ dimensions (Section.\ref{app:Poissons_equation_ND}).

\subsection{Electric forces in 3D} \label{app:force_3D}

In three-dimensions, the electric force between two charges $q_i$ and $q_j$ is calculated as per \textit{Coulomb’s law}:

\begin{equation} \label{eq:coulombs_law_3d}
    \textbf{F} = k \frac{q_i q_j}{r^2} \textbf{e}
\end{equation}

\noindent where $r$ is the distance between the two charges, $k=10^{-7} c^2=\frac{1}{4\pi \varepsilon_0}$ is the \textit{Coulomb constant} (also called the electric force constant or electrostatic constant), with $c$ being the light speed in a vacuum, $\varepsilon_0$ the \textit{vacuum permittivity} introduced later. The unit vector $ \textbf{e}$ implies the direction of the forces - they point in inverse directions as per \textit{Newton's third law}. The force is repulsive for like and the attracting for unlike electric charges. Note that, Eq.\ref{eq:coulombs_law_3d} only holds for three dimensions in which both Coulomb's law and Newton's law of gravity takes the form of distance inverse-squared.

The electric field, e.g. induced by the charge $q_i$, is defined as:

\begin{equation} \label{eq:electric_field}
    \textbf{E}_i = k \frac{q_i}{r^2} \textbf{e}
\end{equation}

\noindent which conveniently simplifies Eq.\ref{eq:coulombs_law_3d} to $\textbf{F}=q_1 \textbf{E}_2$.

Superposition principle applies to the electromagnetic field. The overall electric field and force acting on a particle, induced by others, can be calculated by summing over all the contributions due to individual source particles. Consider a collection of $M$ particles each with charge $q_i$ and distribute in a 3D space. The electric field at arbitrary point (other than the point itself) can be expressed as:

\begin{equation} \label{eq:electric_field_sum}
    \textbf{E}(r) = \frac{1}{4\pi \varepsilon_0} \sum_{i=1}^M \frac{q_i}{r_i^2} \textbf{e}_i
\end{equation}

\noindent where $r_i$ is the distance between a source point and the point of consideration. $\textbf{e}_i$ is a unit vector indicating the direction of the electric field.

The electric field generated by a distribution of charges, as given by the charge density $\rho(\textbf{r})$, can be obtained by replacing the sum by integral:

\begin{equation} \label{eq:electric_field_integral}
    \textbf{E}(r) = \frac{1}{4\pi \varepsilon_0} \iiint \frac{\rho(\textbf{r}_i)}{r_i^2} \textbf{e}_i d^3\textbf{r}
\end{equation}

\noindent which accounts for all contributions from the small charges $\rho(\textbf{r}_i) d^3\textbf{r}$ contained in the infinitesimally small volume element $d^3\textbf{r} \approx dxdydz$.

All these equations are only valid for three dimensions; ideally, we would like to generalize Coulomb's law for \textit{d}-dimension. In the following, we use Poisson's equation (and Laplace's equation) for deriving the electric force in $\textit{d}$ dimensions.

\subsection{The Poisson and Laplace equations for electrostatics} \label{app:force_ND}

We start with the Gauss's law for electricity in differential form:

\begin{equation} \label{eq:Gauss_law}
    \nabla \cdot \textbf{D} = \rho
\end{equation}

\noindent where $\nabla \cdot$ is the divergence operator \footnote{In three-dimensional Cartesian coordinates, for example, the divergence of a continuously differentiable vector field $\textbf{D} = D_x \textbf{i} + D_y \textbf{j} + D_z \textbf{k}$ is defined as the scalar-valued function: $div \textbf{D} = \nabla \cdot \textbf{D} = (\frac{\partial}{\partial x},\frac{\partial}{\partial y},\frac{\partial}{\partial z}) \cdot (D_x,D_y,D_z) = \frac{\partial D_x}{\partial x} + \frac{\partial D_y}{\partial y} + \frac{\partial D_z}{\partial z}$. The divergence operator in cylindrical and spherical coordinates can be similarly derived.}. $\textbf{D}$ is the electric displacement field, $\rho$ is the free-charge density (charges from outside). Gauss's law is one of Maxwell's equations; it relates the distribution of electric charge to the resulting electric field.

Assuming linear, isotropic and homogeneous medium, the following constitutive equation holds:

\begin{equation} \label{eq:linear_field}
    \textbf{D} = \varepsilon \textbf{E}
\end{equation}

\noindent where $\textbf{E}$ is the electric field. $\varepsilon$ is the absolute permittivity \footnote{In electromagnetism, the absolute permittivity $\varepsilon$ is a measure of the electric polarizability of a dielectric material. A material with high permittivity polarizes more in response to an applied electric field than a material with low permittivity, thereby storing more energy in the material.} of the medium. Often, the relative permittivity, which is the ratio of the absolute permittivity $\varepsilon$ and the vacuum permittivity $\varepsilon_0$, i.e. $\varepsilon_r=\varepsilon/\varepsilon_0$, is used. The vacuum permittivity \footnote{It measures how dense of an electric field is "permitted" to form in response to electric charges, and relates the units for electric charge to mechanical quantities such as force given by Coulomb's law.} is the absolute dielectric permittivity of classical vacuum; it is referred to as the \textit{permittivity of free space} or the \textit{electric constant} whose value $\varepsilon_0 \approx 8.8541878128 \times 10^{-12} F/m$ (farads per meter).

The electrostatic force $\textbf{F}$ on a charge $q'$, induced by the electric field $\textbf{E}$, is the product:

\begin{equation} \label{eq:electrostatic_force}
    \textbf{F} = q' \textbf{E}
\end{equation}

Substituting Eq.\ref{eq:linear_field} into Gauss's law in Eq.\ref{eq:Gauss_law}, and assume constant permittivity $\varepsilon$ of the medium, we have:

\begin{equation} \label{eq:Gauss2}
    \nabla \cdot \textbf{E} = \frac{\rho}{\varepsilon}
\end{equation}

In the presence of constant magnetic field or no magnetic field (i.e. if the field is irrotational), we have:

\begin{equation} \label{eq:no_magnetic_field}
    \nabla \times \textbf{E} = 0
\end{equation}

\noindent where $\nabla \times$ is the curl operator. As the curl of any gradient is zero, we can write the electric field $E$ as the \textit{gradient} of a scalar function $\varphi$ (the electric potential):

\begin{equation} \label{eq:electric_potential}
    \textbf{E} = - \nabla \varphi
\end{equation}

\noindent where $\nabla$ here denotes gradient, and $\varphi$ is a real or complex-valued function on a manifold. Eq.\ref{eq:electric_potential} states that, the gradient of the electric potential gives the electric field. The negative sign reflects that the $\varphi$ is the electric potential energy per unit charge.

Substituting Eq.\ref{eq:electric_potential} into Eq.\ref{eq:Gauss2} gives:

\begin{equation} \label{eq:divergence_of_electrical_field}
    \nabla \cdot \textbf{E} = - \nabla^2 \varphi = \frac{\rho}{\varepsilon}
\end{equation}

\noindent where $\nabla \cdot \nabla = \nabla^2 = \Delta$ is the Laplace operator \footnote{Laplace operator is a second-order differential operator in the \textit{n}-dimensional Euclidean space, defined as the divergence ($\nabla \cdot$) of the gradient ($\nabla \varphi$), where $\varphi$ is a twice-differentiable function. Under Cartesian coordinates, the Laplacian of $\varphi$ is thus the sum of all unmixed second partial derivatives: $\Delta \varphi(\textbf{x}) = \sum_{i=1}^d \frac{\partial^2 \varphi(\textbf{x})}{\partial x_i^2}$. Similar expressions can be derived in other coordinates (e.g. cylindrical and spherical coordinates).}. Eq.\ref{eq:divergence_of_electrical_field} relates the divergence of the electric field $\nabla \cdot \textbf{E}$ to the charge density $\rho$ through the negative Laplacian of the potential $\varphi$, which further leads to the \textit{Poisson's equation} for electrostatics:

\begin{equation} \label{eq:poisson_equation}
    \Delta \varphi = \nabla^2 \varphi = - \frac{\rho}{\varepsilon}
\end{equation}

Solving the Poisson's equation Eq.\ref{eq:poisson_equation} gives the electric potential $\varphi$, given the mass density distribution of charges $\rho(\textbf{x})$; however, in order to solve the Poisson's equation, we need to know the charge density distribution $\rho$. Different configurations of $\rho$ gives different solutions. For example, if the charge density $\rho=0$ (i.e. in a region where there are no charges or currents), which gives the \textit{Laplace's equation} (the homogeneous Poisson's equation):

\begin{equation} \label{eq:laplace_equation}
    \nabla^2 \varphi = 0
\end{equation}

\noindent a function satisfying the Laplace equation is called \textit{harmonic} \cite{CristofaroGardiner2017}.

A side note on Poisson's and Laplace equations. Poisson's equation, in its general form as in Eq.\ref{eq:general_3D_poisson_equation}, can be used to describe the diffusion process of a chemical solute or heat transfer, i.e. $\partial \varphi (\textbf{x},t) / \partial t = \kappa \nabla^2 \varphi = \kappa f(\textbf{x,t})$, where $\kappa$ is termed the \textit{diffusivity}. The RHS $f(\textbf{x},t)$ represents external sources (for example, in Eq.\ref{eq:poisson_equation} we have $f(\textbf{x},t)=-\rho/\varepsilon$) which inject inputs into the system (e.g. solute generated by a chemical reaction, or heat provided by external sources). The Laplace equation, as a special case of the Poisson's equation, can be derived as a steady-state condition for the diffusion process (e.g. concentration of chemical solute, heat transfer). The Poisson's equation doesn't have a unique solution, unless extra boundary condition is imposed. For example, if the electrostatic potential $\varphi(\textbf{x})$ is specified on the boundary of a region, then its solution is uniquely determined. Some application scenarios of the Poisson's equation, solutions, as well as uniqueness proofs, can be found in e.g. \cite{Hunt2002}. 

We formerly define $\varphi: \mathbb{R}^d \rightarrow \mathbb{R}$, and its Laplacian, $\Delta \varphi: \mathbb{R}^d \rightarrow \mathbb{R}$, i.e.

\begin{equation} \label{eq:Laplacian_Cartecian}
    \Delta \varphi(\textbf{x}) := \sum_{i=1}^d \partial_{x_i x_i}^2 \varphi(\textbf{x})
\end{equation}

\noindent where $x_i$ is the $i^{th}$ coordinate of $\textbf{x} \in \mathbb{R}^d$. In crafting Eq.\ref{eq:Laplacian_Cartecian}, we have assumed that the Laplacian extends the same form into $d$-dimension.

\subsection{The 3D Poisson's equation} \label{app:Poissons_equation_3D}

Without losing generality \footnote{This subsection mainly follows \cite{Herman2020}.}, we start with solving a 3D \textit{Poisson's equation} generalised from Eq.\ref{eq:poisson_equation}:

\begin{equation} \label{eq:general_3D_poisson_equation}
    \Delta \varphi(\textbf{x}) = \nabla^2 \varphi(\textbf{x}) = f(\textbf{x})
\end{equation}

\noindent where $\varphi: \mathbb{R}^3 \rightarrow \mathbb{R}$ and $f: \mathbb{R}^3 \rightarrow \mathbb{R}$ (the support can be restricted to subset of $\mathbb{R}^3$ though). The RHS $f(x)$ is called the \textit{source} and it is often zero either everywhere or everywhere but some specific region or points. For example, $f(\textbf{x})=c_0 \delta^3(\textbf{x})$, where $c_0$ is a constant, represents a point source. 

We shall apply Fourier transform to Eq.\ref{eq:general_3D_poisson_equation}; before that, we re-cap the 3D Fourier transform:

\begin{equation} \label{eq:3D_fourier_transform}
    \hat{\varphi}(\textbf{k}) = \int \varphi(\textbf{x}) e^{-\textit{i} \textbf{k} \textbf{x}} d^3 \textbf{x}
\end{equation}

\noindent where $d^3 \textbf{x} = dx_1 dx_2 dx_3$, $\textbf{k}=k_{x_1} \textbf{i} + k_{x_2} \textbf{j} + k_{x_3} \textbf{k}$ is a three dimensional wavenumber. The inverse Fourier transform is

\begin{equation} \label{eq:3D_inverse_fourier_transform}
    \varphi(\textbf{x}) = \frac{1}{(2\pi)^3} \int \hat{\varphi}(\textbf{k}) e^{\textit{i} \textbf{k} \textbf{x}} d^3\textbf{k}
\end{equation}

\noindent where $d^3\textbf{k}=dk_{x_1} dk_{x_2} dk_{x_3}$.

Assuming that $\varphi(\textbf{x})$ and its gradients vanish for large $\textbf{x}$, we apply Fourier transform to the generalised 3D Poisson's equation (Eq.\ref{eq:general_3D_poisson_equation}):

\begin{equation} \label{eq:Fourier_transform_of_Laplacian}
    \mathcal{F}[\nabla^2 \varphi(\textbf{x})] = -(k_{x_1}^2 + k_{x_2}^2 + k_{x_3}^2) \hat{\varphi} (\textbf{k}) = \mathcal{F}[f(\textbf{x})] = \hat{f}(\textbf{k})
\end{equation}

\noindent by defining $k^2=k_{x_1}^2 + k_{x_2}^2 + k_{x_3}^2$, we have:

\begin{equation}
    k^2 \hat{\varphi}(\textbf{k}) = \hat{f}(\textbf{k})
\end{equation}

\noindent which gives

\begin{equation}
    \hat{\varphi}(\textbf{k}) = \frac{1}{k^2} \hat{f}(\textbf{k})
\end{equation}

The solution to the generalised 3D Poisson's equation (Eq.\ref{eq:general_3D_poisson_equation}) can be obtained by performing inverse Fourier transform:

\begin{equation} \label{eq:solution_to_3D_generalised_Poissons_equation}
    \varphi (\textbf{x}) = \frac{1}{(2\pi)^3} \int \hat{f}(\textbf{k}) \frac{e^{\textit{i} \textbf{k} \textbf{x}}}{k^2} d^3\textbf{k}
\end{equation}

As a side note, one can also apply inverse Fourier transform to Eq.\ref{eq:Fourier_transform_of_Laplacian}, as a pathway for evaluating the Laplacian $\nabla^2 \varphi(\textbf{x})$, which falls into the method of \textit{spectral derivative}.

As an example demonstrating the use of Eq.\ref{eq:solution_to_3D_generalised_Poissons_equation} to calculate the potential $\varphi(\textbf{x})$, we set $f(\textbf{x}) = - \frac{q}{\varepsilon} \delta(\textbf{x})$, which represents a single point charge $q$ in free space (i.e. a vacuum). The Fourier transform of $f(\textbf{x})$ is:

\begin{equation}
    \hat{f}(\textbf{k}) = \int_{\mathbb{R}^3} [-\frac{q}{\varepsilon} \delta(\textbf{x})] e^{-i\textbf{k}\textbf{x}} d^3\textbf{x} = -\frac{q}{\varepsilon}
\end{equation}

\noindent as the Fourier transform of the delta function $\delta(\textbf{x})$ is 1 over all space.

Substituting $\hat{f}(\textbf{k})$ into Eq.\ref{eq:solution_to_3D_generalised_Poissons_equation}, we obtain:

\begin{equation} \label{eq:Poissons_equation_3D_Fourier_solution}
    \varphi(\textbf{x}) = -\frac{q}{(2\pi)^3 \varepsilon} \int \frac{1}{k^2} e^{i\textbf{k}\textbf{x}} d^3\textbf{k}
\end{equation}

We shall focus on evaluating this term $\int \frac{1}{k^2} e^{i\textbf{k}\textbf{x}} d^3\textbf{k}$. It is more convenient to switch to the spherical coordinates with $d^\textbf{k}=k^2 sin \theta dk d\theta d\phi$. As such, we have:

\begin{equation} \label{eq:Poissons_equation_3D_Fourier_solution1}
    \int \frac{e^{i\textbf{k}\textbf{x}}}{k^2} d^3\textbf{k} = \int_0^{+\infty} \int_{0}^{\pi} \int_0^{2\pi} \frac{e^{i\textbf{k}r cos \theta}}{k^2} k^2 sin \theta d\textbf{k} d\theta d\phi
\end{equation}

\noindent where $r=|\textbf{x}|$ is the radius. We re-arrange the integration sequence in Eq.\ref{eq:Poissons_equation_3D_Fourier_solution1} as:

\begin{equation} \label{eq:Poissons_equation_3D_Fourier_solution2}
    \int \frac{e^{i\textbf{k}\textbf{x}}}{k^2} d^3\textbf{k} = \int_0^{+\infty} \int_0^{2\pi} d\phi \int_{0}^{\pi} e^{i\textbf{k}r cos \theta} sin \theta d\theta d\textbf{k}
    = 2\pi \int_0^{+\infty} \frac{2 sin(\textbf{k}r)}{\textbf{k}r} d\textbf{k} 
    = \frac{2 \pi^2}{r}
\end{equation}

\noindent where we have made use of the integral of \textit{sinc} function over a real line: $\int_{-\infty}^{+\infty} sinc(x)dx = \pi$.
Finally, substituting this result into Eq.\ref{eq:Poissons_equation_3D_Fourier_solution} gives:

\begin{equation} \label{eq:Poissons_equation_3D_Fourier_solution3}
    \varphi(\textbf{x}) = -\frac{q}{(2\pi)^3 \varepsilon} \frac{2 \pi^2}{r} = -\frac{q}{4\pi \varepsilon r}
\end{equation}

\noindent which gives the electric field $\textbf{E}(r) = - \nabla_r \varphi(r) = \frac{q}{4\pi\varepsilon r^2}$, and the resulting force $\textbf{F}(r)=q' \textbf{E}(r)=\frac{q q'}{4\pi\varepsilon r^2}$, which is consistent with the result derived later by integration (Eq.\ref{eq:Coulombs_law_3D}).

\subsection{Solution to the \textit{d}-dimensional Poisson's equation} \label{app:Poissons_equation_ND}

Here we look at the generalised Poisson's equation in \textit{d}-dimensional space. We can repeat the direct and inverse Fourier transforms process as in the 3D case, which gives the solution to the \textit{d}-dimensional, generalised Poisson's equation:

\begin{equation} \label{eq:solution_to_ND_generalised_Poissons_equation}
    \varphi (\textbf{x}) = \frac{1}{(2\pi)^d} \int \hat{f}(\textbf{k}) \frac{e^{\textit{i} \textbf{k} \textbf{x}}}{k^2} d^d\textbf{k}
\end{equation}

\noindent where $\textbf{k}$ is a \textit{d}-dimensional wavenumber, $k^2=\sum_{i=1}^d k_{x_i}^2$, and $d^d\textbf{k}=dk_{x_1} dk_{x_2} ... dk_{x_d}$.

We can also obtain the derivative:
\begin{equation} \label{eq:solution_to_ND_generalised_Poissons_equation_derivative}
    \nabla_{\textbf{x}} \varphi (\textbf{x}) = \frac{1}{(2\pi)^d} \int (i\textbf{k}) \hat{f}(\textbf{k}) \frac{e^{\textit{i} \textbf{k} \textbf{x}}}{k^2} d^d\textbf{k}
\end{equation}

The problem with Eq.\ref{eq:solution_to_ND_generalised_Poissons_equation} (as well as Eq.\ref{eq:solution_to_ND_generalised_Poissons_equation_derivative}) is the intractability of the inverse Fourier integration (which involves multiple integral in spherical coordinates), particularly when $\hat{f}(\textbf{k})$ is complex. Here we take another approach, i.e. using direct integration to solve the Poisson's equation in a general \textit{d}-dimensional space with a point charge. Similar to the 3D example, a single point charge at $\textbf{x}$ can be represented by $\rho(\textbf{x})=q \delta(\textbf{x})$, where $q$ is the total charge. The Poisson's equation for electrostatics (Eq.\ref{eq:poisson_equation}), correspondingly, can be written as:

\begin{equation} \label{eq:poisson_equationsingle_charge}
    \Delta \varphi = \nabla^2 \varphi = - \frac{q}{\varepsilon} \delta(\textbf{x})
\end{equation}

\noindent the solution to this single point charge Poisson's equation is the potential field $\varphi(\textbf{x})$, and therefore the electric field $E(\textbf{x})$ as per their relation Eq.\ref{eq:electric_potential} (and further the force), caused by the given electric charge $q \delta(\textbf{x})$. 

In Cartesian coordinates $(x_1,x_2,...,x_d)$, Eq.\ref{eq:poisson_equationsingle_charge} becomes:

\begin{equation} \label{eq:poisson_equationsingle_charge_Cartesian}
    \sum_{i=1}^d \frac{\partial^2}{\partial x_i^2} \varphi(\textbf{x}) = - \frac{q}{\varepsilon} \delta(\textbf{x})
\end{equation}

Due to the problem's spherical symmetry in \textit{d} dimensions, it is natural to switch to hyper-spherical coordinates. The Laplacian for a radially symmetric function $\varphi(r)$ in hyper-spherical coordinates reduces to:

\begin{equation} \label{eq:hyperspherical_Laplacian}
    \mathcal{L}[\varphi(r)] = \frac{1}{r^{d-1}} \frac{d}{dr} (r^{d-1} \frac{d\varphi}{dr})
\end{equation}

\noindent substituting the Laplacian $\mathcal{L}(\varphi(r)$ into the RHS of the Poisson's equation (Eq.\ref{eq:poisson_equationsingle_charge_Cartesian}) in in hyper-spherical coordinates, we have:

\begin{equation} \label{eq:poisson_equationsingle_charge_hyperspherical}
    \frac{1}{r^{d-1}} \frac{d}{dr} (r^{d-1} \frac{d\varphi}{dr}) = - \frac{q}{\varepsilon} \delta(r)
\end{equation}

Away from the origin (i.e. $r \neq 0$), we have $\delta(r)=0$, which simplifies Eq.\ref{eq:poisson_equationsingle_charge_hyperspherical} to:

\begin{equation}
    \frac{d}{dr} (r^{d-1} \frac{d\varphi}{dr}) = 0
\end{equation}

\noindent integrating it gives:

\begin{equation} \label{eq:away_from_origin_integral}
    r^{d-1} \frac{d\varphi}{dr} = C
\end{equation}

\noindent where $C$ is an integration constant. Solving for $\varphi$, we get:

\begin{equation} \label{eq:solution_to_Poissons_equation}
    \varphi(r) = \frac{C}{(2-d)r^{d-2}} + D
\end{equation}

\noindent again $D$ is another integration constant. We set $D=0$, assuming the potential vanishes at infinity. To determine the constant $C$, we can integrate Eq.\ref{eq:poisson_equationsingle_charge}. First, we integrate the LHS and use the divergence theorem \footnote{The divergence theorem states that the surface integral of a vector field over a closed surface ('the flux through the surface) equals the volume integral of the divergence over the region enclosed by the surface, i.e. \(\int_V \nabla \cdot F dV = \int_S F \cdot \hat{n} dS\), where $S$ is the closed surface surrounding the volume $V$, and $\hat{n}$ is a unit vector directed along the outward normal to $S$. It converts between the surface integral and the volume integral, and generalises to arbitrary dimension. Physically, it means the sum of all sources in a region gives the net flux out of the region.}:

\begin{equation} \label{eq:spherical_integral1}
    \int_V \nabla \cdot (\nabla\varphi) dV = \int_S \nabla\varphi \hat{r} \cdot dS
\end{equation}

\noindent which converts the volume integral into a surface integral, with $S$ being the closed surface surrounding the volume $V$ and $\hat{r}$ a unit vector pointing outward normal to $S$. The quantity on the RHS, i.e. $\int_S \nabla\varphi \hat{r} \cdot dS =  \int_S \textbf{E} \hat{r} dS$ is the electric flux over the surface $S$ (the negative sign from Eq.\ref{eq:electric_potential} has been indicated by the outward pointing unit vector). Due to the spherical symmetry, the integral in Eq.\ref{eq:spherical_integral1} is the product of the surface area of the $d-1$ dimensional hypersphere \footnote{A sphere in a $d$-dimensional space, enclosing a $d$-dimensional ball (called $\textit{d}$-ball, the $d$-1-sphere is the boundary of the $d$-ball), has dimension $d$-1 because it has $d$-1 degrees of freedom due to the fixed radius. For example, the sphere defined by $x_1^2+x_2^2+x_3^2+x_4^2+=r^2$ in 4-dimensional space is called a 3-sphere ($S_3$).} and $\nabla\varphi$, i.e.

\begin{equation} \label{eq:spherical_integral2}
   \int_S \nabla\varphi \hat{r} \cdot dS = S_{d-1} r^{d-1} \times \nabla\varphi
\end{equation}

\noindent in which $S_{d-1}$ is the the surface area of the $d-1$ dimensional \textit{unit} hypersphere, and the surface area of the $d-1$-sphere with radius $r$ is scaled by $r^{d-1}$. $S_{d-1}$ can be obtained \footnote{Alternatively, one can derive the surface area by noting that, the volume of a $\textit{d}$-ball with radius $r$ is: $V_d(r)=r^d V_d(1)=r^d \frac{\pi^{d/2}}{\Gamma(d/2+1)}$ (one can simply verify this in 2D and 3D. NB: if $r$ is fixed, its volume converges to zero as dimension increases, i.e. the volume of a high-dimensional unit ball is concentrated near its surface due to curse of dimensionality), then the surface area of the $\textit{d}$-ball can be derived as the derivative of the volume w.r.t. the radius $r$: $S_{d-1}(r)=\frac{d}{dr} V_d(r) = d r^{d-1} \frac{\pi^{d/2}}{\Gamma(d/2+1)}$. Therefore, the surface area of the $\textit{d}$-ball with unit radius is: $S_{d-1}=d \frac{\pi^{d/2}}{\Gamma(d/2+1)}=\frac{2 \pi^{d/2}}{\Gamma(d/2)}$, which is the same as Eq.\ref{eq:Gaussian_integral3}.} by using the trick of a \textit{d}-dimensional Gaussian integral, which relates this surface area to a known quantity. We know the following:

\begin{equation} \label{eq:Gaussian_integral}
    \int_{-\infty}^{+\infty} e^{-\textbf{x}^2} d^d x = [\int_{-\infty}^{+\infty} e^{-x^2} dx]^d = \pi^{d/2}
\end{equation}

\noindent where we have made use of the known Euler–Poisson integral $\int_{-\infty}^{+\infty} e^{-x^2} dx = \sqrt{\pi}$. This integral can also be related to the surface area of the $d-1$ dimensional \textit{unit} sphere. Writing the LHS of Eq.\ref{eq:Gaussian_integral} in spherical coordinates:

\begin{equation} \label{eq:Gaussian_integral_Gamma1}
     \int_{-\infty}^{+\infty} e^{-\textbf{x}^2} d^d x = S_{d-1} \int_{0}^{+\infty} r^{d-1} e^{-r^2} dr
\end{equation}

\noindent the integral $\int_{0}^{+\infty} r^{d-1} e^{-r^2} dr$ can be evaluated \footnote{Note that, as we are evaluating a definite integral, when $d-1$ is an odd integer, this integral is zero. We can relax this by replacing it with indefinite integral, which gives the same result.} using change of variable, i.e. substituting $u=r^2$ and $du=2rdr$, $dr=du/2r=du/(2\sqrt{u})$ into it, which yields $\int_{0}^{+\infty} r^{d-1} e^{-r^2} dr = \int_{0}^{+\infty} (\sqrt{u})^{d-1} e^{-u} \frac{1}{2\sqrt{u}} du = \frac{1}{2}\int_{0}^{+\infty} u^{\frac{d}{2}-1} e^{-u} du$, which is in the form of the Gamma function $\Gamma(\frac{d}{2})$. Therefore, we have Eq.\ref{eq:Gaussian_integral_Gamma1} reduces to:

\begin{equation} \label{eq:Gaussian_integral_Gamma2}
     \int_{-\infty}^{+\infty} e^{-\textbf{x}^2} d^d x = \frac{1}{2} S_{d-1} \Gamma(\frac{d}{2})
\end{equation}

Equating Eq.\ref{eq:Gaussian_integral} and Eq.\ref{eq:Gaussian_integral_Gamma2}, we arrive at:

\begin{equation} \label{eq:Gaussian_integral2}
     \frac{1}{2} S_{d-1} \Gamma(\frac{d}{2}) = \pi^{d/2}
\end{equation}

\noindent which gives:

\begin{equation} \label{eq:Gaussian_integral3}
     S_{d-1} = \frac{2\pi^{d/2}}{\Gamma(\frac{d}{2})}
\end{equation}

Now we go back to Eq.\ref{eq:spherical_integral2}. We have now $\int_S \nabla\varphi \hat{r} \cdot dS = S_{d-1} r^{d-1} \times \nabla\varphi = \frac{2\pi^{d/2}}{\Gamma(\frac{d}{2})} \times \varphi' = \frac{2\pi^{d/2}}{\Gamma(\frac{d}{2})} r^{d-1} \times C r^{1-d} = \frac{2\pi^{d/2}}{\Gamma(\frac{d}{2})} \times C$ as per Eq.\ref{eq:away_from_origin_integral} (or Eq.\ref{eq:solution_to_Poissons_equation}). The integration of the RHS of Eq.\ref{eq:poisson_equationsingle_charge} is just $-q/\varepsilon$ due to the delta function. Therefore, we have the integration constant $C=-q \Gamma(\frac{d}{2})/(2\pi^{d/2}\varepsilon)$. Finally, the solution to the single point charge Poisson's equation (i.e. Eq.\ref{eq:solution_to_Poissons_equation}) becomes:

\begin{equation} \label{eq:solution_to_Poissons_equation2}
    \varphi(r) = \frac{-q \Gamma(\frac{d}{2})/(2\pi^{d/2}\varepsilon)}{(2-d) r^{d-2}} = \frac{q \Gamma(\frac{d}{2})}{2(d-2)\pi^{d/2}\varepsilon r^{d-2}}
\end{equation}

After obtaining the potential $\varphi$, we can get the electric field $\textbf{E}$ as per Eq.\ref{eq:electric_potential}.

\begin{equation} \label{eq:electric_field_from_potential}
    \textbf{E}_q=-\frac{d \varphi}{dr} = \frac{q \Gamma(\frac{d}{2})}{2\pi^{d/2}\varepsilon r^{d-1}}
\end{equation}

\noindent and the electric force
\footnote{If we are only concerning about the electric force, there is no need to calculate the potential - all we need is the potential derivative $\nabla_{\textbf{x}} \varphi(\textbf{x})$.} exerted on another charge $q'$, by the single point charge $q$ at a distance $r$ from the point charge $q$:

\begin{equation} \label{eq:electric_force_from_potential}
    \textbf{F}_{q \rightarrow q'}= q' \textbf{E}_q = \frac{q q' \Gamma(\frac{d}{2})}{2\pi^{d/2}\varepsilon r^{d-1}}
\end{equation}

\noindent As we are concerning free space, $\varepsilon=\varepsilon_0$ which is the vacuum permittivity. 

\begin{figure}[ht]
    \centering
    \includegraphics[width=0.45\columnwidth]{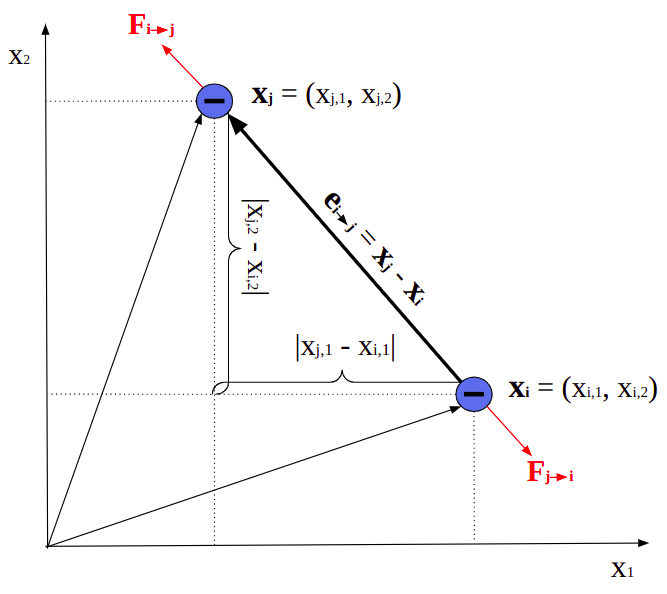}
    \caption{Diagram of repulsive force and distance in a 2D Cartesion coordinate.}
    \label{fig:force_diagram}
\end{figure}

Note that, for notational simplicity, we have dropped all direction notations in our electric potential (Eq.\ref{eq:solution_to_Poissons_equation2}), field (Eq.\ref{eq:electric_field_from_potential}), force (Eq.\ref{eq:electric_force_from_potential}) expressions and their derivations, and focus on their magnitudes. The mindful can add a \textit{unit vector} $\textbf{e}_{q,q'}$, which points away from a positive charge and points to a negative charge, to these formula to indicate the directions of these quantities. In a Cartesian coordinate, for example, if $q$ is at position $\textbf{x}_q$, and $q'$ at position $\textbf{x}_{q'}$, the directional unit vector is then $\textbf{e}_{q \rightarrow q'} := (\textbf{x}_{q'} - \textbf{x}_q)/\lVert \textbf{x}_{q'} - \textbf{x}_q) \rVert_d = (\textbf{x}_{q'} - \textbf{x}_q))/r_{q,q'}$ where $r_{i,j}=\lVert \textbf{x}_{q'} - \textbf{x}_q) \rVert_d$, and $\lVert \cdot \rVert_d$ is the distance measure in $\textit{d}$-dimensional space. By defining the direction of $\textbf{e}_{q \rightarrow q'}$, we have by default assumed repulsive to be positive. Substituting the expression of $r$ into e.g. the force formula gives:

\begin{equation} \tag{\ref{eq:electric_force_from_potential}b}
\begin{aligned}
    \textbf{F}_{q \rightarrow q'} (\textbf{x}_i, \textbf{x}_j) 
    &= \frac{q q' \Gamma(\frac{d}{2})}{2\pi^{d/2}\varepsilon r^{d-1}} \textbf{e}_{q \rightarrow q'} = \frac{q q' \Gamma(\frac{d}{2})}{2\pi^{d/2}\varepsilon r^{d-1}} \frac{\textbf{x}_{q'} - \textbf{x}_q}{r} \\
    &=
    \frac{q q' \Gamma(\frac{d}{2})}{2\pi^{d/2} \varepsilon} \frac{\textbf{x}_{q'} - \textbf{x}_q}{r^d}
    = 
    \frac{q q' \Gamma(\frac{d}{2})}{2\pi^{d/2} \varepsilon} \frac{\textbf{x}_{q'} - \textbf{x}_q}{\lVert \textbf{x}_{q'} - \textbf{x}_q \rVert_2^{d/2}} \\
\end{aligned}
\end{equation}

\noindent where $\lVert \cdot \rVert_2$ denotes the $\textit{L}_2$ norm. A 2D diagram for illustrating the repulsive force and the distance is presented Fig.\ref{fig:force_diagram}.

Eq.\ref{eq:electric_force_from_potential} extends the Coulomb's law for any dimensions. For example, in 1D we have:

\begin{equation} \label{eq:Coulombs_law_1D}
    \textbf{F}_{q,q'} = \frac{q q' \sqrt{\pi}}{2\pi^{1/2}\varepsilon_0 r^{d-1}}= \frac{q q'}{2\varepsilon_0}
\end{equation}

\noindent and for 2D:

\begin{equation} \label{eq:Coulombs_law_2D}
    \textbf{F}_{q,q'} = q' \textbf{E}_q = \frac{q q'}{2\pi \varepsilon_0 r}
\end{equation}

\noindent and for 3D:

\begin{equation} \label{eq:Coulombs_law_3D}
    \textbf{F}_{q,q'} = q' \textbf{E}_q = \frac{q q' \sqrt{\pi}/2}{2\pi^{3/2}\varepsilon_0 r^2} = \frac{q q'}{4\pi\varepsilon_0 r^2}
\end{equation}

\noindent which is the same as Eq.\ref{eq:coulombs_law_3d} and Eq.\ref{eq:Poissons_equation_3D_Fourier_solution3} given:

\begin{equation}
    k = \frac{1}{4\pi \varepsilon_0}
\end{equation}

\section{A note on complexity} \label{app:note_on_complexity}

\begin{figure}[ht]
    \centering
    \includegraphics[width=0.5\columnwidth]{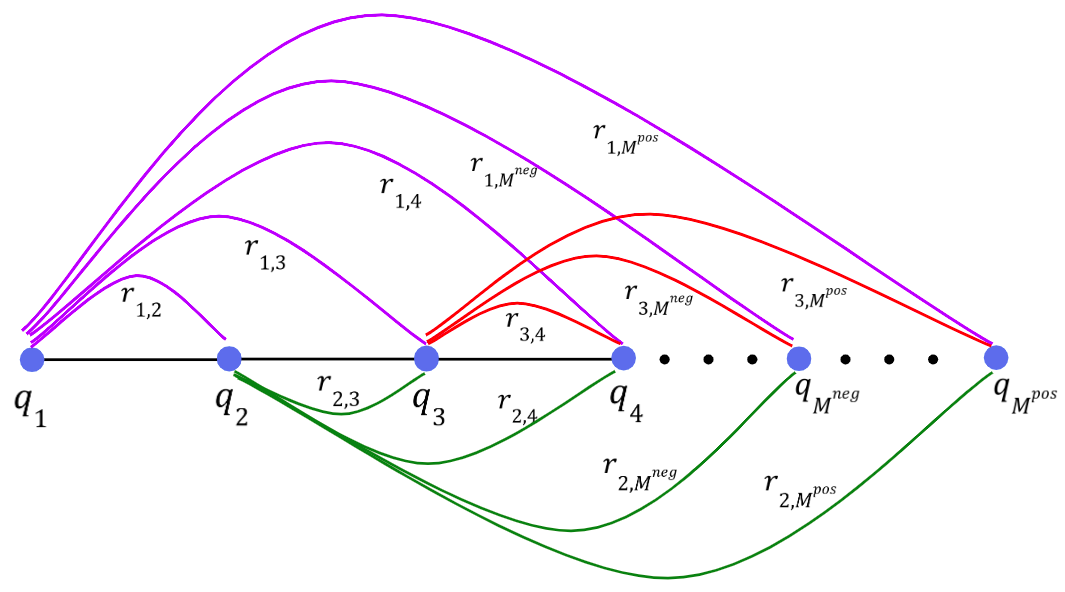}
    \caption{Pairwise distances.}
    \label{fig:pairwise_distance}
\end{figure}

Most computational efforts lie in computing the interacting force between particles. At each iteration, in order to calculate the two forces for a single negative particle $j$, we have to calculate all the pairwise distances $1/r_{i,j}^{d-1}$ and $1/r_{i',j}^{d-1}$ where $i \in \{1,2,...,M^{neg}\}$ and $i' \in \{1,2,...,M^{pos}\}$, as shown in Figure.\ref{fig:pairwise_distance}. That is, we have to evaluate the following distance matrix in each iteration:

\begin{equation} \label{eq:pairwise_distances_matrix}
\begin{aligned}
&R_{(M^{neg}+M^{pos}) \times (M^{neg}+M^{pos})} = \\
&
\begin{tikzpicture}
  \matrix (m) [matrix of math nodes, nodes in empty cells,
    left delimiter={(}, right delimiter={)}, row sep=0.3em, column sep=0.5em]{
    0 & r_{1,2} & \cdots & r_{1,M^{neg}} & \cdots & r_{1,M^{pos}}  \\
    r_{2,1} & 0 & \cdots & r_{2,M^{neg}} & \cdots & r_{2,M^{pos}}  \\
    \vdots & \vdots & \ddots & \vdots & \ddots & \vdots \\
    r_{M^{neg},1} & r_{M^{neg},2} & \cdots & 0 & \cdots & r_{M^{neg},M^{pos}} \\
    \vdots & \vdots & \ddots & \vdots & \ddots & \vdots \\
    r_{M^{pos},1} & r_{M^{pos},2} & \cdots & r_{M^{pos},M^{neg}} & \cdots & 0 \\
  };

  \node[anchor=east] at ([xshift=-2.8em]m-1-1) {$q_{1}$};
  \node[anchor=east] at ([xshift=-2.8em]m-2-1) {$q_{2}$};
  \node[anchor=east] at ([xshift=-2.8em]m-3-1) {$\cdots$};
  \node[anchor=east] at ([xshift=-2.8em]m-4-1) {$q_{M^{neg}}$};
  \node[anchor=east] at ([xshift=-2.8em]m-5-1) {$\cdots$};
  \node[anchor=east] at ([xshift=-2.8em]m-6-1) {$q_{M^{pos}}$};

  \node[anchor=south] at ([yshift=1.0em]m-1-1) {$q_{1}$};
  \node[anchor=south] at ([yshift=1.0em]m-1-2) {$q_{2}$};
  \node[anchor=south] at ([yshift=1.0em]m-1-3) {$\cdots$};
  \node[anchor=south] at ([yshift=1.0em]m-1-4) {$q_{M^{neg}}$};
  \node[anchor=south] at ([yshift=1.0em]m-1-5) {$\cdots$};
  \node[anchor=south] at ([yshift=1.0em]m-1-6) {$q_{M^{pos}}$};
\end{tikzpicture}
\end{aligned}
\end{equation}

\noindent which can be pre-calculated or calculated when needed. Calculating the whole matrix $R$ takes order $\mathcal{O}((M^{neg}+M^{pos})^2)$. The computational efforts will be huge if either $M^{neg}$ or $M^{pos}$ is large. Of course, half of the computation can be saved by making use of the symmetry $r_{i,j}=r_{j,i}$.

Let's take a closer look at the repulsive force in Eq.\ref{eq:repulsive_force2} by expanding it in matrix form:

\begin{equation} \label{eq:repulsive_force2_matrixExpansion}
\tag{cc.\ref{eq:repulsive_force2}}
\begin{aligned}
\textbf{F}_j^{rep} (\textbf{x}_j)
&=
\frac{\Gamma(\frac{d}{2})}{2\pi^{d/2}\varepsilon_0}
\textbf{q}_j
\mathbf{\Lambda}
\textbf{q}_{1:M^{neg}}
\\
&=
\frac{\Gamma(\frac{d}{2})}{2\pi^{d/2}\varepsilon_0}
\begin{pmatrix}
q_j & q_j & \cdots & q_j
\end{pmatrix}
\begin{pmatrix}
1/r_{1,j}^{d-1} & 0 & \cdots & 0 \\
0 & 1/r_{2,j}^{d-1} & \cdots & 0 \\
\vdots & \vdots & \ddots & \vdots \\
0 & 0 & \cdots & 1/r_{i,j}^{d-1}
\end{pmatrix}
\begin{pmatrix}
q_1 \\
q_2 \\
\vdots \\
q_{M^{neg}}
\end{pmatrix}
\end{aligned}
\end{equation}

\noindent similar matrix expansion can be written with the attracting force in Eq.\ref{eq:attracting_force} for particle $j$. 

When calculating $\textbf{F}_j^{rep} (\textbf{x}_j)$ and $\textbf{F}_j^{attr} (\textbf{x}_j)$ for the particle $j$, we need to retrieve the relevant pairwise distance vector, i.e. the row $r_{j,\cdot}$ or column $r_{\cdot,j}$, from the distance matrix $R$ in Eq.\ref{eq:pairwise_distances_matrix}. However, if necessary (i.e. if either $M^{neg}$ or $M^{pos}$ is large), we can take a neighbourhood search to identify a neighbourhood set and calculate only the pairwise distances between particles within the neighbourhood set. This shrinks the distance vector for particle $j$. 

\section{Mapping concepts in electrostatics and probability}

Quantities in electrostatics can be mapped to the concepts in probability distributions as follows:

\begin{table}[ht]
    \centering
    \renewcommand{\arraystretch}{1.5}  
    \begin{tabular}{c|c|c|c}
    \textbf{Output Type} & \textbf{Electrostatics} & \textbf{Probability} & Used in this work \\
    \hline
    Scalar-valued & $\varphi(\textbf{x})$ & $\log p(\textbf{x})$ & - \\
    \hline
    \multirow{3}{*}{Vector-valued} & \multirow{3}{*}{\shortstack{$\nabla \varphi(\textbf{x})$ \\ $\textbf{E} = -\nabla \varphi(\textbf{x})$ \\ $\textbf{F} = q' \textbf{E} = -q' \nabla \varphi(\textbf{x})$}} & \multirow{3}{*}{$\nabla \log p(\textbf{x})$} & \multirow{3}{*}{\textbf{F} $\propto$ $p(\textbf{x})$ or $\log p(\textbf{x})$} \\
    & & & \\
    & & & \\
    \hline
    Scalar-valued & $\Delta \varphi(\textbf{x}) = \nabla \cdot \nabla \varphi(\textbf{x})$ & $\text{tr}(\nabla^2 \log p(\textbf{x}))$ & - \\
    \end{tabular}
    \caption{Correspondence of concepts in electrostatics and probability.}
    \label{tab:electrostatics_probability_correspondence}
\end{table}

For example, if we assume $\varphi(\textbf{x}) = k \nabla \log p(\textbf{x})$, then $\textbf{F}=-q\textbf{E}=-q \nabla \varphi(\textbf{x}) = -q k \nabla \log p(\textbf{x})$, which is similar to the driving force term $k(\cdot,\cdot) \nabla \log p(\textbf{x})$ in SVGD. Therefore, in order to calculate the force $\textbf{F}$, we only need to gain knowledge about $\nabla \varphi(x)$, which says, equivalently in probability, that we only need to know $\nabla \log p(\textbf{x})$, the gradient field can be conveniently estimated by techniques such as \textit{score matching} \cite{hyvarinen2005sm}.

\section{Toy examples: more details} \label{app:toy_examples}

\paragraph{Gaussian densities} The two target Gaussian densities are as follows:
\\
Unimodal:
\[ 
p(\textbf{x})=\mathcal{N}(\mathbf{\mu},\Sigma) 
\]
with $\mathbf{\mu}=(0.5,0.5)$ and $\Sigma=0.05\textit{I}$.
\\
Bimodal: 
\[
p(\textbf{x})=0.7\mathcal{N}(\mathbf{\mu}_1,\Sigma_1) + 0.3\mathcal{N}(\mathbf{\mu}_2,\Sigma_2) 
\] with $\mathbf{\mu}_1=(0,0)$, $\Sigma_1=
    \begin{bmatrix}
    1 & -0.5 \\
    -0.5 & 1
    \end{bmatrix}
    $, $\mathbf{\mu}_2=(4,4)$, $\Sigma_2=
    \begin{bmatrix}
    1 & 0.5 \\ 
    0.5 & 1
    \end{bmatrix}$.

In inferring the unimodal Gaussian, we draw negative charges from the uniform prior: $p^0(\textbf{x})=U(0,0.5) \times U(0,0.5)$; In inferring the bimodal Gaussian, we draw negative charges from the uniform prior: $p^0(\textbf{x})=U(-3,7) \times U(-3,7)$.

\paragraph{Other three densities} The other 4 un-normalised target densities used are:
\\
Moon-shared \cite{haario1999adaptive,Wang2021EVI}:
\[
\rho(x) \propto \exp \left\{ -\frac{x_1^2}{2} - \frac{1}{2} \left(10 x_2 + 3 x_1^2 - 3 \right)^2 \right\}
\]
\\
Double-banana \cite{rezende2016variational,liu2019understanding}:
\[
\rho(x) \propto \exp \left\{ -2 \left( (x_1^2 + x_2^2 - 3)^2 \right) + \log \left( e^{-2(x_1-2)^2} + e^{-2(x_2+2)^2} \right) \right\}
\]
\\
Wave density \cite{rezende2016variational,chen2018unified,Wang2021EVI}:
\[
\rho(x) \propto \exp \left\{ -\frac{1}{2} \left[ \frac{x_2 - \sin\left(\frac{\pi x_1}{2}\right)}{0.4} \right]^2 \right\}
\]

\paragraph{Implementation} 
The mesh grid for all 3 other densities is within $[-3,3] \times [-3,3]$; for the bi-modal Gaussian it is $[-3,7] \times [-3,7]$; for uni-modal Gaussian it is $[0,1] \times [0,1]$. The 400 negative charges are initialised with 2D standard Gaussian for the moon-shaped case, and uniformly for all other cases (uniform within the whole mesh grid, except for the uni-modal Gaussian case where the initialisation area is $[0,0.5] \times [0,0.5]$). 

Euler updating rule (Eq.\ref{eq:linear_position_update}) is used and $M^{pos} = 50 \times 50, M^{neg}=400, \sigma_{noise}=0$ applied. The overall forces across negative charges are normalised in each iteration before calculating the displacement. The charge magnitudes $q^{neg}=q=1.0$. Permittivity of free space $\varepsilon_0 = 8.854e-12$. Euler proportionality constant $\tau = 0.1$. Particles are evolved for $T=100$ iterations. For all three densities, at equilibrium, we discard the particles outside the grid area.

\section{Neal's funnel: implementation details} \label{app:neals_funnel_implementation_details}

\paragraph{The $MMD^2$ metric} is defined as \cite{Arbel2019,Wang2021EVI}:

\begin{equation}
\text{MMD}^2 = \frac{1}{N^2} \sum_{i,j=1}^{N} k(x_i, x_j) + \frac{1}{M^2} \sum_{i,j=1}^{M} k(y_i, y_j) 
- \frac{2}{NM} \sum_{i=1}^{N} \sum_{j=1}^{M} k(x_i, y_j)
\end{equation}

\noindent with the polynomial kernel $k(x, y) = \left( \frac{x^\top y}{3} + 1 \right)^3$.

\paragraph{HMC} The \textit{PyMC} Python API \cite{pymc2023} is used to implement the HMC method. In total, 1 chain with 4000 samples is drawn, with 1000 burn-up samples discarded. These 4000 samples are then thinned evenly to produce 400 samples.

\paragraph{MH} Initialised from near zero, we use a standard normal distribution as the proposal distribution, the simplified acceptance probability is defined as the ratio of the new sample to current position. 

\paragraph{LMC} We initialise 400 samples uniformly within $[-7,3] \times [-7,3]$, and evolve them through 10000 iterations; in each iteration, each particle is evolved through the \textit{Langevin dynamics} \cite{wang2019stein}:

\begin{align*}
\mathbf{x}_{t+1} &= \mathbf{x}_t + \epsilon_t \nabla \log p(\mathbf{x}_t) + \sqrt{2\epsilon_t} \mathbf{z}_t \\
\epsilon_t &= a (b + t)^{-c} \\
\mathbf{z}_t &\sim \mathcal{N}(0, I)
\end{align*}

\noindent with $a=0.01, b=1, c=0.55$ and 10000 iterations for each sample. $t$ is the iteration number. 

We also records the step time, mean negative log-likelihood (NLL), and squared maximum mean discrepancy squared (MMD\(^2\)) in each iteration.

\paragraph{SVGD} SVGD inference updates particle positions as per \cite{Liu2016SVGD}:
\[
\textbf{x}_i^{l+1} = \textbf{x}_i^l + \epsilon \phi^*_{q,p}(\textbf{x}_i^l)
\]
and 
\[
\phi^*(\textbf{x}_i^l) = \frac{1}{M} \sum_{j=1}^M [k(\textbf{x}_i^l,\textbf{x}_j^l)\nabla_{\textbf{x}} \log p(\textbf{x}_j^l) + \nabla_{\textbf{x}_j} k(\textbf{x}_i^l,\textbf{x}_j^l)]
\]
We referred to the implementation by \cite{Wang2021EVI}, in which a RBF kernel $\exp(\lVert x-x'\rVert/2h^2)$ is used with the hyper-parameter length-scale $h$=median pairwise distance. The step size used is 0.01. The same 400 uniformly initialised samples (as used in LMC) are evolved, with total number of iterations 100,000. We also records the step time, mean NLL, and MMD\(^2\) in each iteration.

\paragraph{EVI} We refer to the implementation by \cite{Wang2021EVI}, in which the RBF kernel with length-scale $h=0.1$ is used. The step size used is $1.0 \times 10^-7$. The same 400 uniformly initialised samples (as used in LMC) are evolved, with total number of iterations 2,500,000. We also records the step time, mean NLL, and MMD\(^2\) in each iteration.

\paragraph{EParVI} The 400 negative charges are initialised uniformly, the same as that of LMC, SVGD and EVI. Euler updating rule (Eq.\ref{eq:linear_position_update}) is used and $M^{pos} = 10,000, M^{neg}=400, \sigma_{noise}=0$ applied. The overall forces across negative charges are normalised in each iteration before calculating the displacement. The charge magnitudes $q^{neg}=q=1.0$. Permittivity of free space $\varepsilon_0 = 8.854e-12$. Euler proportionality constant $\tau = 0.1$. Particles are evolved for $T=100$ iterations. The $MMD^2$ and $NLL$ metrics are evaluated based on particles within the mesh grid.

\section{Bayesian logistic regression: more details} \label{app:BLR}

\paragraph{Posterior}

The likelihood can be written as:

\[
p(\boldsymbol{y}|\textbf{x}, \boldsymbol{\omega}) = \prod_{i=1}^N \left[ \frac{1}{1 + \exp(-\boldsymbol{\omega}^T \textbf{x}_i)} \right]^{y_i} \left[ 1 - \frac{1}{1 + \exp(-\boldsymbol{\omega}^T \textbf{x}_i)} \right]^{1 - y_i}
\]
\noindent combining with the prior:

\[
p(\boldsymbol{\omega}) = \mathcal{N}(\boldsymbol{\omega}; \boldsymbol{0}, \alpha \boldsymbol{I}) = \frac{1}{(2\pi \alpha)^{d/2}} \exp \left( -\frac{1}{2\alpha} \boldsymbol{\omega}^T \boldsymbol{\omega} \right)
\]

\noindent where $d=4$ is the number of features (not considering intercept), we obtain the coefficients $\boldsymbol{\omega}$ posterior:

\[
p(\boldsymbol{\omega}|\boldsymbol{y}, \textbf{x}) \propto \prod_{t=1}^N \left[ \frac{1}{1 + \exp(-\boldsymbol{\omega}^T \textbf{x}_i)} \right]^{y_i} \left[ 1 - \frac{1}{1 + \exp(-\boldsymbol{\omega}^T \textbf{x}_i)} \right]^{1 - y_i} \cdot \exp \left( -\frac{1}{2\alpha} \boldsymbol{\omega}^T \boldsymbol{\omega} \right)
\]

Taking log we have:

\[
\log p(\boldsymbol{\omega}|\boldsymbol{y}, \textbf{x}) = \sum_{t=1}^N \left[ y_i \log \sigma(\boldsymbol{\omega}^T \textbf{x}_i) + (1 - y_i) \log (1 - \sigma(\boldsymbol{\omega}^T \textbf{x}_i)) \right] - \frac{1}{2\alpha} \boldsymbol{\omega}^T \boldsymbol{\omega} + \text{constant}
\]

\noindent where $\sigma(x) = \frac{1}{1 + \exp(-x)}$ is the logistic function. One can then conveniently take gradient w.r.t $\omega$ to use gradient-based sampling algorithms such as HMC.

\paragraph{Gradient of posterior}

\noindent The gradient of \(\log \sigma(\omega^T \textbf{x}_i)\) w.r.t. \(\omega\) is:
\[ \nabla_\omega \log \sigma(\omega^T \textbf{x}_i) = (1 - \sigma(\omega^T \textbf{x}_i)) \textbf{x}_i \]

\noindent The gradient of \(\log (1 - \sigma(\omega^T \textbf{x}_i))\) w.r.t. \(\omega\) is:
\[ \nabla_\omega \log (1 - \sigma(\omega^T \textbf{x}_i)) = -\sigma(\omega^T \textbf{x}_i) \textbf{x}_i \]

\noindent Combining these, the gradient of the likelihood term is:
\[ \sum_{i=1}^N \left[ y_i (1 - \sigma(\omega^T \textbf{x}_i)) \textbf{x}_i - (1 - y_i) \sigma(\omega^T \textbf{x}_i) \textbf{x}_i \right] \]

\noindent which simplifies to:
\[ \sum_{i=1}^N \left[ (y_i - \sigma(\omega^T \textbf{x}_i)) \textbf{x}_i \right] \]

\noindent On the other hand, the gradient of the prior term w.r.t. \(\omega\) is:
\[ -\frac{1}{\alpha} \omega \]

\noindent Combining the gradients of the likelihood and the prior, we get:
\[ \nabla_\omega \log p(\omega | y, X) = \sum_{t=1}^N \left[ (y_i - \sigma(\omega^T \textbf{x}_i)) \textbf{x}_i \right] - \frac{1}{\alpha} \omega \]

\noindent In matrix form, this can be written as:
\[ \nabla_\omega \log p(\omega | y, \textbf{X}) = \textbf{X}^T (y - \sigma(\textbf{X} \omega)) - \frac{1}{\alpha} \omega \]

\noindent where \(\sigma(\textbf{X} \omega)\) is the vector of sigmoid values for each row of \(\textbf{X}\).

\paragraph{Implementation} The Iris dataset contains 150 observations, each instance with 4 features. The whole dataset is 70$\%$:30$\%$ split into training and test sets, with training contains 105 instances and test has 45 instances. Each observation is represented by the feature-label pair ($\textbf{x}_i,y_i$). Features are normalised before being used.

The SVGD, EVI and EParVI methods use the same number of particles (i.e. 400), and share the same initialisation. Specifically:

\paragraph{HMC} In total, 1 chain with 2000 samples is drawn. After discarding the 1000 burn-up samples, the first 400 samples are used.

\paragraph{SVGD}  A RBF kernel $\exp(\lVert x-x'\rVert/2h^2)$ is used with the hyper-parameter length-scale $h$=median pairwise distance. The step size used is 0.01. Total number of iterations is 200,000. 

\paragraph{EVI}  A RBF kernel with length-scale $h=0.1$ is used. The step size used is $1\times 10^-7$. Total number of iterations is $2.5 \times 10^8$. 

\paragraph{EParVI} Euler updating rule (Eq.\ref{eq:linear_position_update}) is used and $M^{pos} = 12^4, M^{neg}=400, \sigma_{noise}=0$ applied. Note the region spans $[-3,3] \times [-3,3]$, and the granularity (i.e. resolution) is therefore 0.5. The overall forces across negative charges are normalised in each iteration before calculating the displacement. The charge magnitudes $q^{neg}=1.0$ and $q=1.0 \times 10^7$. The relatively large value of $q$ is due to small chain product of the un-normalised likelihood in Eq.\ref{eq:BLR_target}; to make it on the same scale as the repulsive force, by trial and error we need to set a large value for $q$. a strategy to avoid small (absolute) values of the likelihood is to take log, as we did in inferring the LV dynamical system parameters in Section.\ref{sec:LV_systm}. Euler proportionality constant $\tau = 0.1$. Particles are evolved for $T=60$ iterations.

\paragraph{Implementation tricks} For complex geometries, more steps and more charges are required. Updating the particle positions using different rules may lead to different results. Interestingly, in our experiments we find that, if we update the negative positions according to Eq.\ref{eq:nonlinear_position_update2}, in the limit all negative charges are absorbed to high probability grid points. Also, if the $\Delta t$ used in the Verlet or damped Verlet methods is too large, then the particle system is unstable and cannot converge to a steady-state.

One can add random noise to perturb the negative charge positions to avoid overlapping at grid points and to avoid getting trapped in local optima. One can choose to discard particles falling out of the grids (e.g. those get trapped in low-density regions) at the end or sequentially. This also saves computational efforts. Without an MH step, charges can take large, risky steps and jump outside the region. \cite{hanson2005halftoning} also addressed that, appropriate conditions need to be specified at the boundary of the region. In our experiments, we designed an annealing scheme for reducing $q^{pos}$ over iterations to promote within-mode diversity.

It is generally suggested to pre-evaluate the relative magnitudes of repulsive and attracting forces and scale $q^{pos}$ properly to ensure they are on the same scale; otherwise, e.g. when repulsive forces is of order larger than attracting force, the density center will never attract the particles. Effect of the ratio $q/q_i$ is important: a small ratio can make the repulsive force dominate and lead to explosion; a large value can accelerate convergence, but also be too restrictive - the final positions of negative charges overlap and diversity reduces. 

\section{LV dynamical system: more details} \label{app:LV_system}

\paragraph{Posterior} The logarithm of the posterior (Eq.\ref{Eq:LV_posterior}), with priors written in Eq.\ref{Eq:LV_prior} and likelihood in Eq.\ref{Eq:LV_likelihood_func}, is:

\begin{multline} \label{Eq:LV_log_posterior} 
    \log p(\theta|X',Y') = \log p(a,b,c,d|X',Y') = \log p(a) + \log p(b) + \log p(c) + \log p(d) + \log p(X',Y'|a,b,c,d) = \\ 
    \sum_{i=a,b,c,d}\log(\frac{1}{max_i-min_i})
    -2N\log(\sqrt{2\pi}\sigma) - \sum_{i=1}^N (\log x_i' + \log y_i') - \frac{1}{2\sigma^2} \sum_{i=1}^N [(\log x_i' - \log x_i)^2 + (\log y_i' - \log y_i)^2]
\end{multline}

\noindent where $min_a=min_b=min_c=min_d=0.001, max_a=max_b=max_c=max_d=1$ as specified by Eq.\ref{Eq:LV_prior}, and noise level $\sigma=0.25$. Note that, though not explicitly shown, the ODE solutions $(x_i,y_i)$ are functions of $\theta=(a,b,c,d)$.

To make all log posteriors positive, we use the following offset log posterior as target for EParVI:

\[
\log p(\theta|X',Y') - \log p(\theta_0|X',Y') = \log \frac{p(\theta|X',Y')}{p(\theta_0|X',Y')}
\]

\noindent where $p(\theta_0|X',Y')=\min_{\theta \in \mathcal{G}} \log p(\theta|X',Y')$, $\mathcal{G}$ is the set of all grid points (locations of positive charges). Alternatively, $p(\theta_0|X',Y')$ can be chosen to be a value smaller than the log posterior evaluated at any point in the space of consideration. 

\begin{figure}[ht] 
    \centering
    \begin{subfigure}[b]{0.50\columnwidth}
        \includegraphics[width=\linewidth]{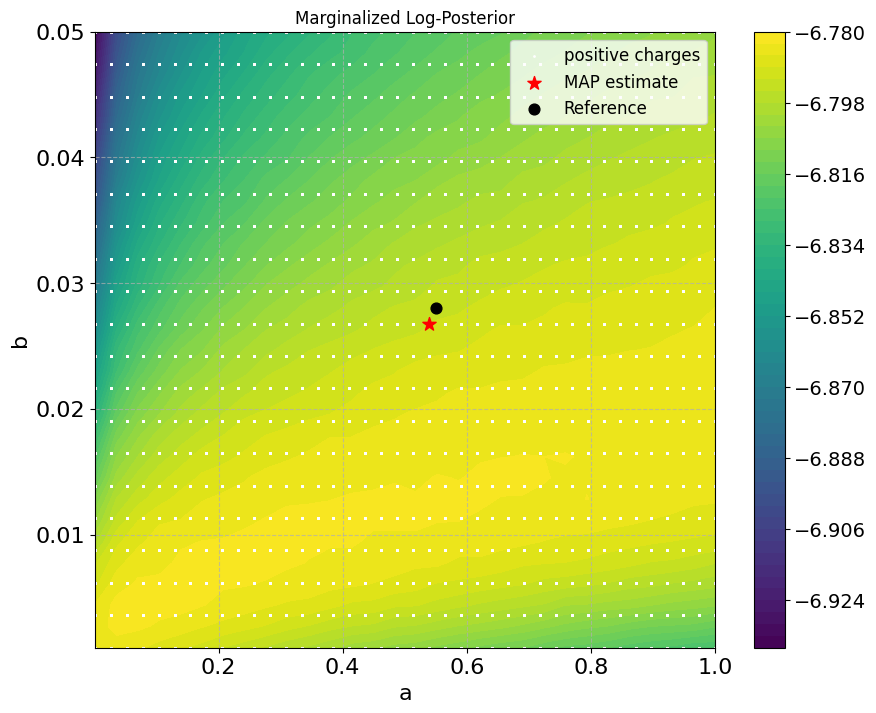}
    \end{subfigure}
    \hfill 
    \begin{subfigure}[b]{0.48\columnwidth}
        \includegraphics[width=\linewidth]{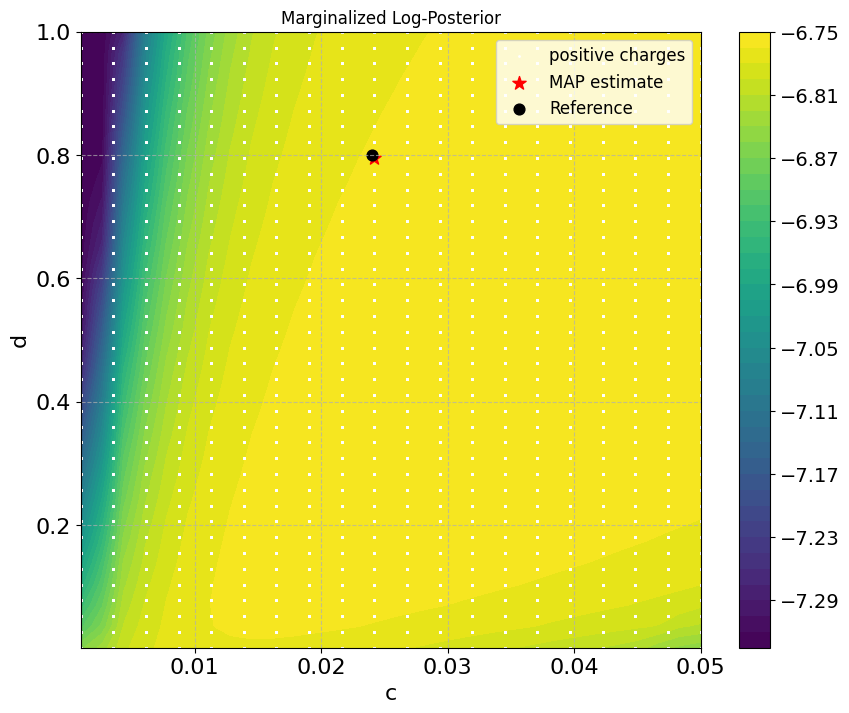}
    \end{subfigure}
    \caption{Marginal log posteriors. Left: marginalising out $c,d$; right: marginalising out $a,b$.}
    \label{fig:LV_marginal_log_posterior}
\end{figure}

By querying the log posterior formula, we obtain the log posterior values at 810000 equidistant points (30 along each dimension). Marginalising out two dimensions, we arrive at the theoretical, 2D marginal log posteriors as presented in Fig.\ref{fig:LV_marginal_log_posterior}.

As the score $\nabla_{\theta} \log p(\theta|X',Y')$ is used in gradient-based samplers such as LMC and HMC, we derive it as follows \footnote{More details can be found in the author's coming book "Some empirical observations on Stein variational gradient descent".}:

\begin{equation} \label{Eq:LV_gradient_log_posterior}
\begin{aligned}
    & \frac{\partial \log p(a,b,c,d | X',Y')}{\partial a} = \frac{1}{\sigma^2} [(\frac{\log X' - \log X}{X})^T \frac{\partial X}{\partial a} + (\frac{\log Y' - \log Y}{Y})^T \frac{\partial Y}{\partial a}] \\ 
    & \frac{\partial \log p(a,b,c,d | X',Y')}{\partial b} = \frac{1}{\sigma^2} [(\frac{\log X' - \log X}{X})^T \frac{\partial X}{\partial b} + (\frac{\log Y' - \log Y}{Y})^T \frac{\partial Y}{\partial b}] \\
    & \frac{\partial \log p(a,b,c,d | X',Y')}{\partial c} = \frac{1}{\sigma^2} [(\frac{\log X' - \log X}{X})^T \frac{\partial X}{\partial c} + (\frac{\log Y' - \log Y}{Y})^T \frac{\partial Y}{\partial c}] \\
    & \frac{\partial \log p(a,b,c,d | X',Y')}{\partial d} = \frac{1}{\sigma^2} [(\frac{\log X' - \log X}{X})^T \frac{\partial X}{\partial d} + (\frac{\log Y' - \log Y}{Y})^T \frac{\partial Y}{\partial d}] \\
\end{aligned}
\end{equation}

\noindent where the posterior with uniform priors in Eq.\ref{Eq:LV_log_posterior} is used. The parameter sensitivities $(\frac{\partial X}{\partial a},\frac{\partial X}{\partial b},\frac{\partial X}{\partial c},\frac{\partial X}{\partial d})$ can be solved by differentiating the LV dynamics in Eq.\ref{Eq:LV_dynamics}.

\begin{figure}[H]
    \centering
    \begin{subfigure}[b]{0.35\textwidth}
        \centering
        \includegraphics[width=\textwidth]{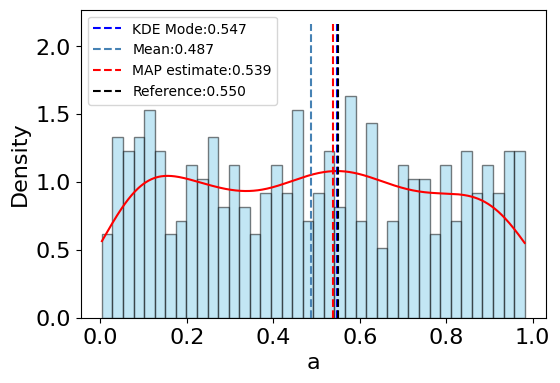}
        \caption{a}
    \end{subfigure}
    \begin{subfigure}[b]{0.35\textwidth}
        \centering
        \includegraphics[width=\textwidth]{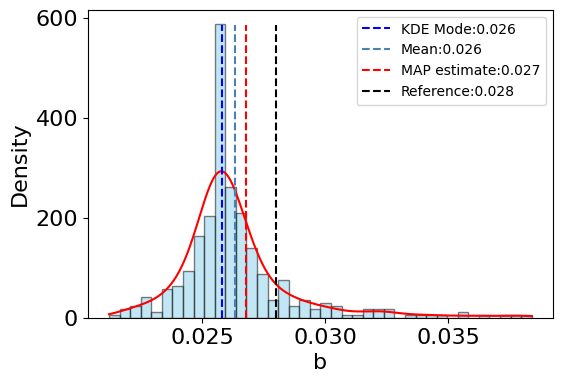}
        \caption{b}
    \end{subfigure}
    \\
    \begin{subfigure}[b]{0.35\textwidth}
        \centering
        \includegraphics[width=\textwidth]{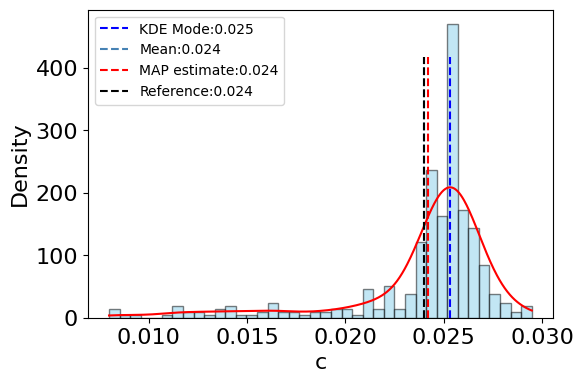}
        \caption{c}
    \end{subfigure}
    \begin{subfigure}[b]{0.35\textwidth}
        \centering
        \includegraphics[width=\textwidth]{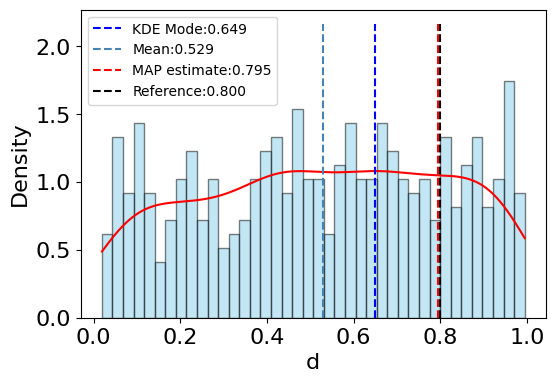}
        \caption{d}
    \end{subfigure}
    \caption{Empirical marginal distributions for $a,b,c,d$.}
    \label{fig:LV_empirical_marginal_dist}
\end{figure}

\paragraph{Implementation and results} Implementation details and supplementary results are listed as follows:

\paragraph{EParVI} The uniform prior space spans $[0.001,1] \times [0.001,0.05] \times [0.001,0.05] \times [0.001,1]$. Each dimension has 40,20,20,40 grid points, respectively. Total number of positive charges $M^{pos}=40\times20\times20\times40=640000$. Total number of negative charges $M^{neg}=400$. Positive charge magnitudes $q_j=q \log p(\theta_j|X',Y')$ are pre-calculated before all iterations, with maximum magnitude $q=10^{-5}$. Noise level $\sigma_{noise}=0$. Euler proportionality constant $\tau = 0.01$. Particles are evolved for $T=82$ iterations. Each iteration takes $\sim 2700$ seconds to run. Overall forces across negative charges are normalised in each iteration before Euler updates. Total runtime for 82 iterations is $\sim 221400$ seconds. Particle positions marginally change after 50 iterations. The empirical, marginal distribution for each dimension, calculated from the 400 particles, is pictured in Fig.\ref{fig:LV_empirical_marginal_dist}.

\begin{figure}[H]
    \centering
    \begin{subfigure}[b]{0.50\columnwidth}
        \includegraphics[width=\linewidth]{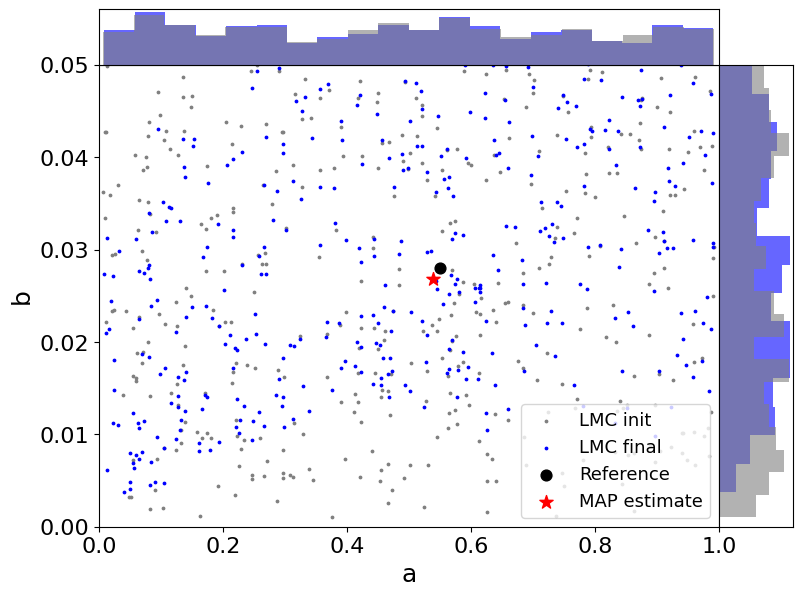}
    \end{subfigure}
    \hfill 
    \begin{subfigure}[b]{0.48\columnwidth}
        \includegraphics[width=\linewidth]{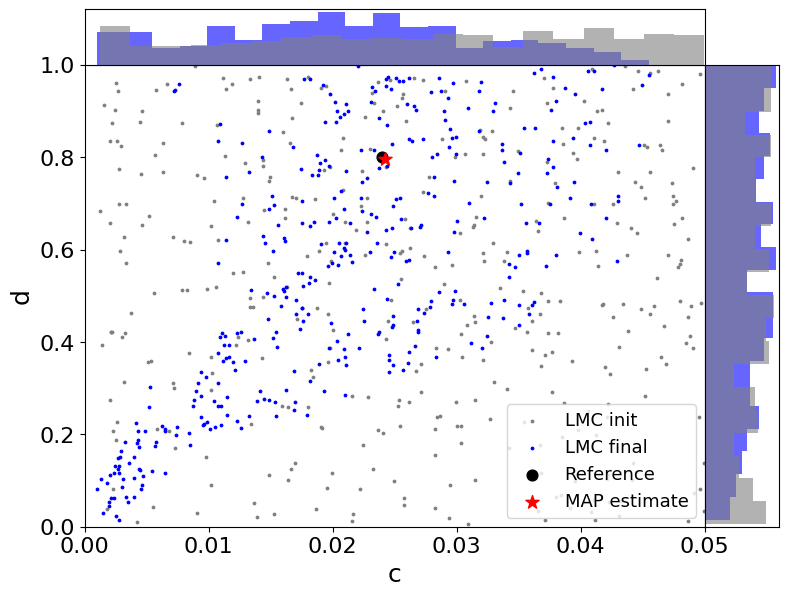}
    \end{subfigure}
    \caption{EPaVI: initial and final particle distributions.}
    \label{fig:LV_prior_and_posterior_LMC}
\end{figure}

\paragraph{LMC} We use similar LMC implementation as described in Appendix.\ref{app:neals_funnel_implementation_details}, with $a=1e-8, b=1, c=0.55$. 400 $\textbf{x}_{0}$ are initialised uniformly (same initial samples used in EParVI), and use Langevin dynamics to trace each sample through 10000 steps. Total runtime is 24178 seconds. The final positions of the 400 samples are presented in Fig.\ref{fig:LV_prior_and_posterior_LMC}. The final shapes are somewhat consistent with the theoretical snapshots in Fig.\ref{fig:LV_log_posterior}, with much dispersion.

\paragraph{HMC} The No-U-Turn sampler (NUTS) \cite{NUTS2014}, an advanced HMC implementation, is used with PyMC3 library \cite{pymc2023}. The author crafted the specific ODEOp and solver. Two chains, each with 1900 samples, are drawn with target acceptance rate 0.8. Total runtime is 366 seconds. The first chain is used, with 1500 burn-in samples discarded. Initial conditions used are $x_0=33.956, y_0=5.933$. Note that, the likelihood used for HMC is $x'(t_i) \sim \mathcal{N}(x(t_i),(e^{0.25})^2)$ and $y'(t_i) \sim \mathcal{N}(y(t_i),(e^{0.25})^2)$, which is different from that of EParVI specified in Eq.\ref{eq:LV_likelihood}.

\begin{figure}[H]
    \centering
    \includegraphics[width=0.65\linewidth]{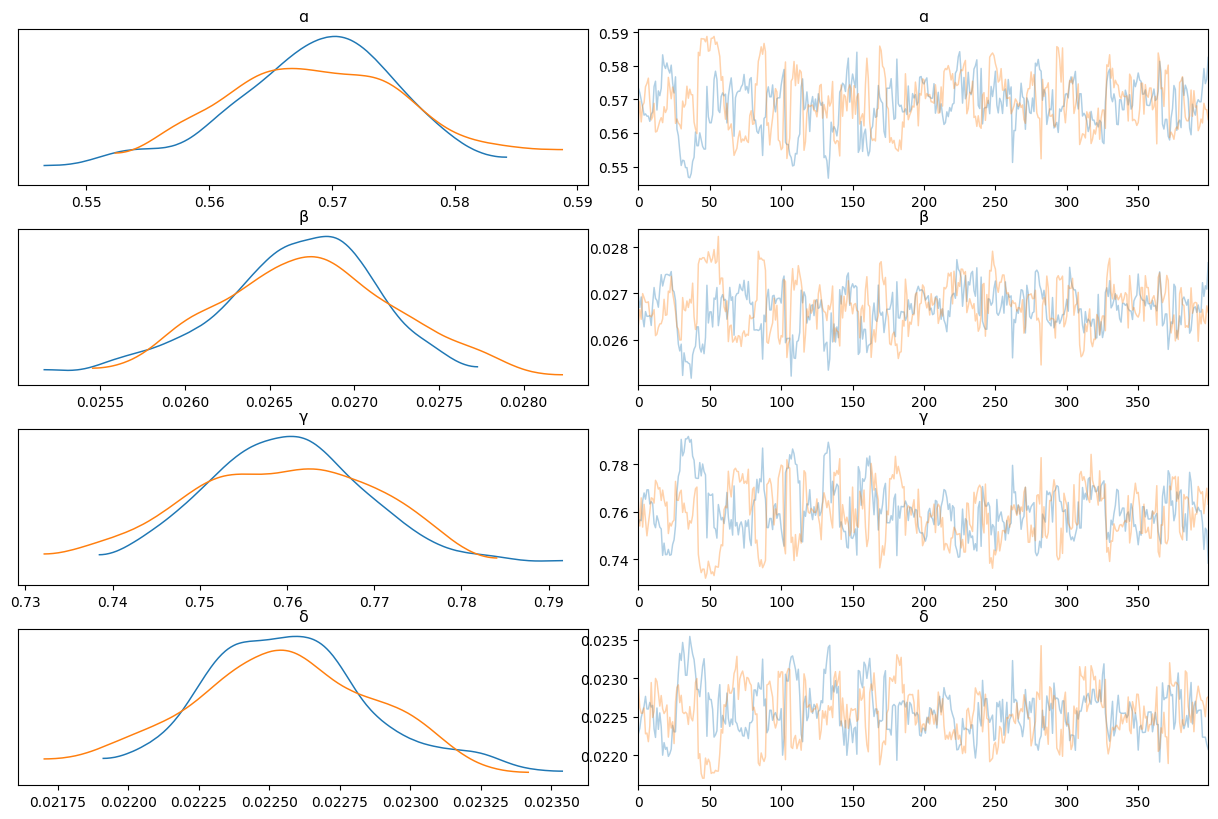}
    \caption{HMC samples. In the figure, $\alpha=a$, $\beta=b$, $\delta=c$, \\$\gamma=d$.}
    \label{fig:LV_HMC_two_chains}
\end{figure}

\begin{figure}[H]
    \centering
    \subfloat[\centering $a$]{{\includegraphics[width=0.65\columnwidth]{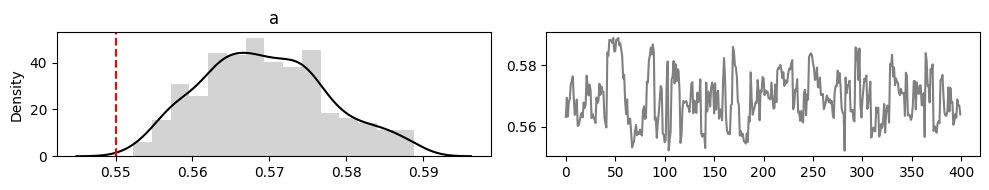}}}
    \\
    \subfloat[\centering $b$]{{\includegraphics[width=0.65\columnwidth]{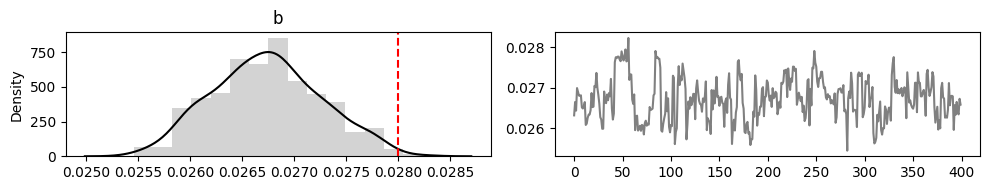}}}
    \\
    \subfloat[\centering $c$]{{\includegraphics[width=0.65\columnwidth]{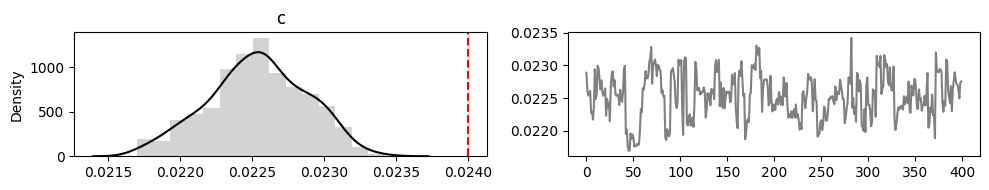}}}
    \\
    \subfloat[\centering $d$]{{\includegraphics[width=0.65\columnwidth]{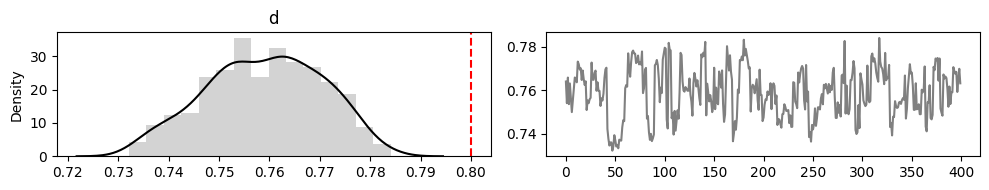}}}
    \caption{\textit{Lotka-Volterra} system identification: HMC inference of $\theta=[a,b,c,d]$ using the real-world data as shown in Fig.\ref{fig:LV_data}. Red dashed vertical line is the reference value from \cite{LV_Carpenter}.}
    \label{fig:LV_HMC_chain1}
\end{figure}

\begin{figure}[H]
    \centering
    \includegraphics[width=1.0\linewidth]{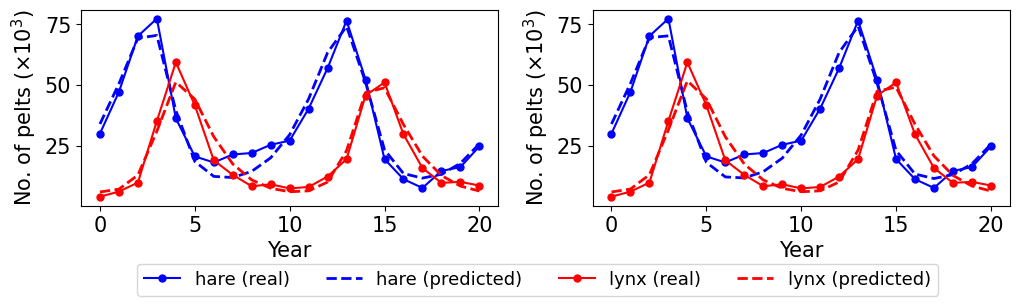}
    \caption{HMC inferred posterior predictives. Left: MAP-based predictions. Right: mean-based predictions.}
    \label{fig:LV_posterior_predictive_HMC}
\end{figure}

The two chains are presented in Fig.\ref{fig:LV_HMC_two_chains}, with left showing the sample histogram based KDE density and on the right sample trajectories. We observe that the two chains agree on the estimations, and the samples, although small numbers, concentrate around the the means and modes. We extract one chain (yellow one in Fig.\ref{fig:LV_HMC_two_chains}) and take a closer look in Fig.\ref{fig:LV_HMC_chain1} at the sample estimates, we observe that, given our HMC model, the results are very close to the reference values (note the small horizontal scales). These empirically estimated values are summarised in the third row of Table.\ref{tab:LV_inference}. We then make predictions using the empirically estimated posterior means and MAPs based on this single chain, as shown in Fig.\ref{fig:LV_posterior_predictive_HMC}. Again, we observe good match with the real world recorded data.

\paragraph{SVGD} Total number of initial particles: $M$=2000, overall number of iterations: $L=100$, kernel: RBF ($2h^2$=$med^2/\log M$ where $med$ is the median distance between particles), learning rate: $\epsilon$=$0.001/\Vert \phi^*(\boldsymbol z_i^l) \Vert_2$, where $i$ is the particle index, $l=1,2,...,L$ is the iteration number. Total runtime time: 8662 seconds. Valid number of particles after 100 iterations: 1454.

\paragraph{EVI} EVI is not used due to our intractability of negative states encountered in simulating the LV dynamics when calculating the gradients.

\section{Further discussions} \label{app:other_discussions}
Here we discuss more aspects and prospects related to the method we proposed.

\paragraph{Hyperparameters} Hyper-parameters used in the SPH method include: number of positive charges (number of mesh grid points) $M^{pos}$, number of negative charges $M^{neg}$, negative charge magnitude (normally set to be 1.0), maximum positive charge magnitude $q$, total number of iterations $T$, position updating parameters $\tau$, $\tau'$ and $\Delta t$, noise level $\sigma$, and others (e.g. annealing parameters). 

EParVI doesn't have many parameters to tune - most of them can be empirically specified. Number of positive charges $M^{pos}$ depends on the granularity to be achieved, it relies on dimension of the problem $m$, and it has great impact on the inference accuracy. It is set to be $M^{pos}=M^m$ if regular, equidistant mesh grid is used. $M^{neg}$ can impact the inference quality as well, in general it can be much smaller than $M^{pos}$, and there is no hard requirement on it. the maximum positive charge magnitude $q$ can be conveniently chosen by trial and error, one can choose it to make repulsive and attracting forces are on the same order in a trial step. $T$ should be large enough to ensure a transient or steady state which is close to the target (one can monitor some metrics such as iteration-wise density or clustering metric to decide when to terminate). Small values for $\Delta t$, $\tau$ and $\tau'$ are required; adaptive schemes can be designated, i.e. in regions where the attracting force is large (i.e. high density regimes), smaller move is expected. One can also use different values for different dimensions (otherwise one dimension may converge fast while other dimensions not). For example, if the parameters to be inferred are of different scales, e.g. $a,d \in [0,1]$ while $b,c \in [0,0.05]$ in the LV system, we can use different Euler proportionality constant $\tau$ and different step sizes $\Delta t$ and damping constant $\tau'$ for $a,d$ and $b,c$. The noise level $\sigma$ can perturb particle positions to avoid getting trapped in local optimal configuration (Eq.\ref{eq:loca_optima}); it is normally set to be small as compared to the move. There are other implementation tricks, as listed in Appendix.\ref{app:BLR}, to facilitate the inference procedure.

\paragraph{Effect of mesh grid} It is reported by \cite{Schmaltz2010} that, hexagonal structure is the optimal structure in 2D for achieving equilibrium; however, this cannot be well represented on the rectangular grid. If the rectangular grid is to be used, unpleasant artefacts such as multiple particles cluster at the same grid position can be resulted. This can be solved by constraining the particles to concentrate on moving towards nearest grid point, which can be realised by adding an extra, distance-inverse force component to the force acting on each particle (Eq.1s) which regularizes large movement from nearby grid points. 

\paragraph{A temporal graph network perspective} Dynamic movement of a group of interacting particles can be viewed as the evolution of a connected graph network over time. One can use network analysis techniques to identify the important contributing edges for each node at at any cross-section of time.

\paragraph{Insights from molecular dynamics} Highly accurate and efficient first-principle based molecular simulations provide insights as well as a wealth of training data for modelling IPS \cite{Duignan2024}. For example, accurate quantum chemistry algorithms, e.g. density functional theory (DFT), have made it possible to build fast and highly accurate models of water \cite{Bore2023}. Neural network potentials (NNPs\cite{Duignan2024}), for example, learn the interaction between molecules based on quantum chemistry data and is capable of predicting the atomic potential field in a molecular system. It can capture the complex interaction patterns between molecules with high computational efficiency ($\mathcal{O}(M)$) and accuracy. These insights can be brought to create analogy when devising a particle system.

\paragraph{Sample clustering for multi-modal inference} In the post-sampling stage, one may be interested in empirically estimating the statistics of $p(\textbf{x})$ for use of e.g. \textit{posterior predictive} in Bayesian modelling. Summarising the statistics of a distribution is not easy, particularly when it's high-dimensional and multi-modal. With the (quasi) samples drawn, we can perform e.g. clustering to group the samples and summarise the statistics in each cluster. This can be achieved using available clustering and partition algorithms such as \textit{k-d-tree}. 

\paragraph{Learning $p(x)$ or $\log p(x)$} In some cases, directly learning $p(x)$ can be numerically unstable, especially when $p(x)$ is very small or large (e.g. chain products of likelihoods). Working with $\log p(x)$ helps to mitigate these issues because the log-transformation compresses the range of values, leading to more stable numerical computation. Also, the gradients of the log-probability $\nabla \log p(x)$ are more stable and easier to handle numerically than the gradients of p(x) itself, and therefore used commonly in ParVI (e.g. SVGD). However, for ParVI methods, if one decide to performance inference on $\log p(x)$, then a density estimator is needed to recover $p(x)=e^{\log p(x)}$. This is the case with Section.\ref{sec:LV_systm}.

\paragraph{Change of variables}
For vector-valued random variable $x'$ and $x$ who live in the same dimension $\mathbb{R}^m$, the \textit{change of variables} principle applies:

\begin{equation} \label{eq:change_of_variables}
    p(x) = p'(x') / |det J(x')| = p'(x') / | det(\frac{\partial x}{ \partial x'}) | = p'(x') / | det[\frac{\partial f_\theta(x')}{ \partial x'}] | \text{ with } x = f(x')
\end{equation}

\noindent where $|det[\frac{\partial f_\theta(x')}{ \partial x'}]|$ is the absolute value of the determinant of the Jacobian matrix $J{f_\theta}(x_i), i=1,2,...M$. If a density is transformed multiple times, the chain rule of product can be used to arrive at the final transformed pdf, yielding a \textit{normalizing flow} \footnote{Normalizing flows \cite{NormalisingFlows2015} allow for learning a complex transformation between probability distributions.}. The EParVI inference algorithm for change of variables is presented in Algo.\ref{algo:change_of_variables}. Note that, It requires computing the determinant of the Jacobian matrix of the transform, which can be computationally expensive for high dimensions. 

\begin{algorithm}[ht]
\caption{EParVI inference for change of variables}
\label{algo:change_of_variables}
\begin{itemize}
\item{\textbf{Inputs}: target probability density function $p(x)$ with $x \in \mathbb{R}^m$; probability density function $p'(x')$ with $x' \in \mathbb{R}^{m}$. Number of particles $M$.}  
\item{\textbf{Outputs}: Particles $\{\textbf{x}_i \in \mathbb{R}^m\}_{i=1}^{M}$ whose empirical distribution approximates the target density $p(\textbf{x})$.} 
\end{itemize}
\vskip 0.06in
1. \textit{Generate samples in low dimensions}. Use EParVI to generate sample particles $x'_i \sim p'(x')$, $i=1,2,...M$. \\

2. \textit{Neural normalizing flow training}. Initialize a neural network $f_{\theta}$: $\mathbb{R}^{m} \rightarrow \mathbb{R}^{m}$ which accepts $x'$ and outputs $x$.
    \begin{addmargin}[1em]{0em}%
    (1) For each particle $i$, compute $x_i$ = $f_{\theta}(x'_i)$ and individual loss $\mathcal{L}_i = -\log p(x_i) - \log |\det J{f_\theta}(x'_i)|$. \\ 
    (2) Compute mean loss $\frac{1}{M} \sum_{i=1}^M \mathcal{L}_i$. Back-propagating the error to update network weights. \\
    \end{addmargin}

2. \textit{Inference and original sample reconstruction}. Mapping $x'_i$ to $x_i$.
    \begin{addmargin}[1em]{0em}%
    (1) For each particle $x'_i, i=1,2,...,M$, project it using the trained neural network: $x_i=f_{\theta}(x'_i)$. \\
    (2) Perform density estimation (e.g. KDE) to obtain $\hat{p}'(x'_i)$. \\
    (3) Estimate target density $\hat{p}(x_i)=\hat{p}'(x'_i) / |det J{f_\theta}(x'_i)|$. \\
    \end{addmargin}

3. \textit{Return} $x_i, i=1,2,...,M$ and the estimated density $\hat{p}(x)$. \\
\end{algorithm}

\paragraph{Generalise to high dimensions}

Eq.\ref{eq:change_of_variables} requires a bijective and differentiable map $x=f_\theta (x')$. If we have the low dimensional samples $x'_i \in \mathbb{R}^{m'}$, e.g. produced by EParVI, and we want to project them into higher dimensional samples $x_i \in \mathbb{R}^{m}$, with $m'<m, i=1,2,...M$. That is, we first generate samples $x' \sim p'(x')$ in low dimensions, which is feasible using EParVI, and then project the low-dimensional samples into high dimensions via $x=f_\theta(x')$, obtaining $x \sim p(x)$ and preserving the probability structure. The change of variables principle can no longer be applied in this case (the Jacobian matrix is non-square). Instead, we employ a neural network to represent this non-linear mapping $x=f_\theta(x')$, with the training objective \footnote{Note that, regularization term can be added to promote sample diversity and smooth mapping.} $\mathcal{L}(x'_i) = -\log p(x_i) - \mathcal{L}'_i$, where $\mathcal{L}'_i$ is a measure of the contribution of the projected sample $x_i$ to the diversity of $p(x)$ in the projected space $\mathbb{R}^m$, ensuring the mapping properly captures the structure of the target pdf $p(x)$. The algorithm is presented in Algo.\ref{algo:low_to_high_dimension_inference}. 

\begin{algorithm}[ht]
\caption{Projected EParVI inference}
\label{algo:low_to_high_dimension_inference}
\begin{itemize}
\item{\textbf{Inputs}: target probability density function $p(x)$ with $x \in \mathbb{R}^m$; probability density function $p'(x')$ with $x' \in \mathbb{R}^{m'}$. Number of particles $M$.}  
\item{\textbf{Outputs}: Particles $\{\textbf{x}_i \in \mathbb{R}^m\}_{i=1}^{M}$ whose empirical distribution approximates the target density $p(\textbf{x})$.} 
\end{itemize}
\vskip 0.06in
1. \textit{Generate samples in low dimensions}. Use EParVI to generate sample particles $x'_i \sim p'(x')$, $i=1,2,...M$. \\

2. \textit{Neural network training}. Initialize a neural network $f_{\theta}$: $\mathbb{R}^{m'} \rightarrow \mathbb{R}^{m}$ which accepts $x'$ and outputs $x$.
    \begin{addmargin}[1em]{0em}%
    (1) For each particle $i$, compute $x_i$ = $f_{\theta}(x'_i)$ and individual loss $\mathcal{L}i = -\log p(x_i) - \mathcal{L}'_i$. \\ 
    (2) Compute mean loss $\frac{1}{M} \sum_{i=1}^M \mathcal{L}_i$. Back-propagating the error to update network weights. \\
    \end{addmargin}

2. \textit{Inference and original sample reconstruction}. Mapping $x'_i \in \mathbb{R}^{m'}$ to $x_i \in\mathbb{R}^m$.
    \begin{addmargin}[1em]{0em}%
    (1) For each particle $x'_i, i=1,2,...,M$, project it using the trained neural network: $x_i=f_{\theta}(x'_i)$. \\
    (2) Perform density estimation (e.g. KDE) to obtain $\hat{p}(x)$. \\
    \end{addmargin}

3. \textit{Return} $x_i, i=1,2,...,M$ and the estimated density $\hat{p}(x)$. \\
\end{algorithm}

Algo.\ref{algo:low_to_high_dimension_inference} provides a way to estimate high-dimensional distributions by working in a lower-dimensional space, which can be more computationally efficient. Quality of the final estimation depends on the expressiveness of the neural network, design of the loss function and the training process. The diversity loss $\mathcal{L}'_i$ can be designed using strategies for example:
(1) Product of singular values of the Jacobian. This gives a measure of the 'volume change':
$\mathcal{L}'_i = \sum_{j=1}^{m'} \log \sigma_j[J{f_\theta}(x'_i)]$, where $\sigma_j$ are the singular values of the Jacobian $J{f_\theta}$. This approach accounts for expansion in all directions, and it is computationally expensive (routinely, SVD is performed to obtain the sigular values). One can use only the largest singular value which gives an upper bound on the expansion.
(2) Project the Jacobian onto an orthonormal basis in the higher-dimensional space: $\mathcal{L}'_i = \log |\det[J{f_\theta}(x'i)^T J{f_\theta}(x'_i)]|^{1/2}$.
(3) Use the trace of the Gram matrix of the Jacobian as a measure of magnitude: $\mathcal{L}'_i = \log \text{Tr}[J{f_\theta}(x'i)^T J{f_\theta}(x'_i)]$.
(4) Use the Frobenius norm of the Jacobian as a simple scalar measure: $\mathcal{L}'_i = \log |J{f_\theta}(x'_i)|_F$. Each of these loss designs provides a way to capture the transformation's properties, by quantifying the 'expansion' or 'contraction' of the transformation without requiring a square Jacobian.

\end{document}